\documentclass{article}

\newif\ifdvcneurips
\dvcneuripsfalse

\usepackage[margin=1in]{geometry}
\usepackage{natbib}
\usepackage{times}
\usepackage[utf8]{inputenc}
\usepackage[T1]{fontenc}
\usepackage{url}
\usepackage{booktabs}
\usepackage{amsfonts}
\usepackage{amsmath}
\usepackage{amssymb}
\usepackage{microtype}
\usepackage{graphicx}
\usepackage{algorithm}
\usepackage{algorithmic}
\usepackage{amsthm}
\usepackage{tikz}
\usetikzlibrary{positioning,arrows.meta,shapes.geometric,decorations.pathmorphing,calc}
\usepackage[section]{placeins}  
\usepackage[colorlinks=true,linkcolor=blue,citecolor=blue,urlcolor=blue]{hyperref}

\theoremstyle{plain}
\newtheorem{proposition}{Proposition}
\theoremstyle{remark}
\newtheorem*{remark}{Remark}

\newcommand{\R}{\mathbb{R}}
\newcommand{\E}{\mathbb{E}}
\newcommand{\bx}{\mathbf{x}}
\newcommand{\bu}{\mathbf{u}}
\newcommand{\bz}{\mathbf{z}}

\DeclareMathOperator{\MI}{MI}

\DeclareMathOperator{\TC}{TC}

\newcommand{\dvcCodeFootnote}{\footnote{Code, benchmark configurations, and reproduction scripts are available at \url{https://github.com/KempnerInstitute/DVC}.}}
\newcommand{\dvcCodeAssetSource}{\url{https://github.com/KempnerInstitute/DVC}}

\title{Dynamic Vine Copulas: Detecting and Quantifying\\Time-Varying Higher-Order Interactions}
\author{%
\begin{tabular}[t]{c@{\qquad\qquad}c}
Houman Safaai\textsuperscript{*} & Alessandro Marin Vargas\\
Kempner Institute at Harvard University & Stanford University
\end{tabular}
}
\date{}

\begin{document}
\maketitle
\begingroup
\renewcommand{\thefootnote}{*}
\footnotetext{Correspondence: \texttt{houman\_safaai@harvard.edu}.}
\endgroup

\begin{abstract}
Time-varying dependence is often modeled with dynamic correlations or Gaussian graphical models, but multivariate systems can change through tail behavior, asymmetry, or conditional structure even when correlations are nearly stable. We introduce \textbf{Dynamic Vine Copulas (DVC)}, a temporal vine-copula framework for estimating and diagnosing sequence-wide non-Gaussian dependence. DVC fixes a chosen vine factorization for comparability; the framework applies to C-, D-, and R-vines, and our experiments use fixed-root-order C-vines. Pair-copula states evolve through smooth parameter trajectories or temporally regularized family-switching paths. The main diagnostic is a held-out comparison between a full vine and its matched 1-truncated version, which separates flexible first-tree pairwise dependence from evidence contributed by higher-tree conditional terms. At the population level, under a correct fixed vine and the simplifying assumption, this contrast equals the higher-tree component of a vine total-correlation decomposition; in finite samples, it is a predictive diagnostic. In controlled benchmarks, DVC detects Student-$t$ degrees-of-freedom changes, Clayton-to-Gumbel switches, and recurrent conditional-interaction episodes missed or conflated by Gaussian dynamic baselines. The higher-tree score remains near zero in pairwise-only regimes and rises during conditional-interaction regimes. On Allen Visual Behavior Neuropixels data, DVC identifies a reproducible time-indexed higher-tree signal that is positive across held-out splits and vanishes under a decorrelated null, indicating simultaneous cross-area dependence. DVC therefore provides a flexible temporal copula model and an interpretable test of whether temporal dependence changes are pairwise or conditional.
\end{abstract}
  
  \section{Introduction}
  
  Dependence in complex systems changes over time, but not always in ways visible to correlation.
  In cortical circuits, noise correlations vary with attention, task engagement, and brain state \citep{Cohen2011}, and weak pairwise correlations can still produce coordinated population states \citep{Schneidman2006}; whether higher-order interactions beyond pairwise structure are needed for population coding remains an active question \ifdvcneurips\else\citep{Ganmor2011,Panzeri2022}\fi.
  In financial markets, crises can alter tail co-movement and asymmetric dependence beyond ordinary correlation \citep{LonginSolnik2001,Embrechts2002,Patton2006IER}; in behavioral and ecological systems, interaction structure can reorganize across contexts.
  In these settings, the scientific question is not only whether dependence changes, but \emph{how}: through Gaussian pairwise correlation, non-Gaussian tail behavior, asymmetric co-fluctuation, or conditional structure that appears only after accounting for other variables.
  Figure~\ref{fig:overview} previews the vine decomposition, temporal coupling, and held-out diagnostic used below.
  
  \begin{figure}[!t]
  \centering
  \resizebox{0.99\linewidth}{!}{%
  \begin{tikzpicture}[
    font=\small,
    node distance=6mm,
    every node/.style={font=\small},
    varnode/.style={circle, draw, thick, minimum size=5mm, inner sep=0pt, fill=blue!10},
    edge1/.style={draw, thick, color=black!80},
    edge2/.style={draw, thick, color=orange!75!black, dashed},
    edge3/.style={draw, thick, color=red!70!black, dotted},
    dynnode/.style={circle, draw, thick, minimum size=3.6mm, inner sep=0pt, fill=blue!7, font=\tiny},
    weakedge/.style={draw, color=gray!50, line width=0.35pt},
    pairdyn/.style={draw, color=blue!75!black, line width=1.5pt},
    conddyn/.style={draw, color=orange!80!black, dashed, line width=1.45pt},
    hodyn/.style={draw, color=red!75!black, dotted, line width=1.55pt},
    panel/.style={font=\bfseries\small},
    arrowstyle/.style={-{Latex[length=2mm]}, thick},
  ]
  
  \node[panel, anchor=west] at (-0.3, 3.0) {(A) C-vine decomposition};
  
  \node[varnode] (v1) at (0.6, 1.6) {$X_1$};
  \node[varnode] (v2) at (-0.4, 0.6) {$X_2$};
  \node[varnode] (v3) at (1.6, 0.6) {$X_3$};
  \node[varnode] (v4) at (0.6, -0.4) {$X_4$};
  
  \draw[edge1] (v1) -- node[above left=-1pt, font=\scriptsize] {$c_{12}$} (v2);
  \draw[edge1] (v1) -- node[above right=-1pt, font=\scriptsize] {$c_{13}$} (v3);
  \draw[edge1] (v1) -- node[right=-1pt, font=\scriptsize] {$c_{14}$} (v4);
  
  \node[font=\scriptsize, anchor=north] at (0.6, -0.55) {$T_1$: pairwise};
  
  \node[varnode, scale=0.65, fill=orange!15] (e12) at (3.1, 1.3) {\tiny $12$};
  \node[varnode, scale=0.65, fill=orange!15] (e13) at (4.1, 1.3) {\tiny $13$};
  \node[varnode, scale=0.65, fill=orange!15] (e14) at (3.6, 0.3) {\tiny $14$};
  \draw[edge2] (e12) -- node[above, font=\scriptsize] {$c_{23|1}$} (e13);
  \draw[edge2] (e13) -- node[right=-1pt, font=\scriptsize] {$c_{34|1}$} (e14);
  \node[font=\scriptsize, anchor=north] at (3.6, -0.55) {$T_2$: conditional};
  
  \node[varnode, scale=0.55, fill=red!15] (e123) at (5.6, 1.0) {\tiny $23|1$};
  \node[varnode, scale=0.55, fill=red!15] (e134) at (6.6, 1.0) {\tiny $34|1$};
  \draw[edge3] (e123) -- node[above, font=\scriptsize] {$c_{24|13}$} (e134);
  \node[font=\scriptsize, anchor=north] at (6.1, -0.55) {$T_3$: higher};
  
  \node[panel, anchor=west] at (8.25, 3.0) {(B) Fixed vine, changing edges};
  
  \begin{scope}[xshift=8.25cm, yshift=-0.32cm]
    \node[draw, rounded corners=1pt, fill=yellow!15, align=center, font=\scriptsize, inner sep=2pt, minimum width=37mm]
      (state) at (2.05, 2.72) {one vine factorization across time\\[-1pt]
      edge states vary with $t$};
    \draw[->, thick] (0.05, 1.72) -- (4.15, 1.72) node[right, font=\scriptsize] {$t$};
    \foreach \x/\lab in {0.45/$t_1$,2.0/$t_2$,3.55/$t_3$} {
      \draw[gray!60] (\x, 1.62) -- (\x, 1.82);
      \node[font=\scriptsize, anchor=south] at (\x, 1.83) {\lab};
    }
  
    \begin{scope}[xshift=0.45cm, yshift=0.54cm]
      \node[dynnode] (a1) at (0, 0.85) {$1$};
      \node[dynnode] (a2) at (-0.48, 0.20) {$2$};
      \node[dynnode] (a3) at (0.48, 0.20) {$3$};
      \node[dynnode] (a4) at (0, -0.43) {$4$};
      \draw[pairdyn] (a1) -- (a2);
      \draw[weakedge] (a1) -- (a3);
      \draw[weakedge] (a1) -- (a4);
      \draw[weakedge, dashed] (a2) -- (a3);
      \draw[weakedge, dotted] (a2) to[bend right=22] (a4);
      \node[font=\tiny, align=center, anchor=north] at (0, -0.62) {$T_1$: pair\\$c_{12}$ Gauss};
    \end{scope}
  
    \begin{scope}[xshift=2.0cm, yshift=0.54cm]
      \node[dynnode] (b1) at (0, 0.85) {$1$};
      \node[dynnode] (b2) at (-0.48, 0.20) {$2$};
      \node[dynnode] (b3) at (0.48, 0.20) {$3$};
      \node[dynnode] (b4) at (0, -0.43) {$4$};
      \draw[weakedge] (b1) -- (b2);
      \draw[weakedge] (b1) -- (b3);
      \draw[weakedge] (b1) -- (b4);
      \draw[conddyn] (b2) -- (b3);
      \draw[weakedge, dotted] (b2) to[bend right=22] (b4);
      \node[font=\tiny, align=center, anchor=north] at (0, -0.62) {$T_2$: conditional\\$c_{23|1}$ Clayton};
    \end{scope}
  
    \begin{scope}[xshift=3.55cm, yshift=0.54cm]
      \node[dynnode] (c1) at (0, 0.85) {$1$};
      \node[dynnode] (c2) at (-0.48, 0.20) {$2$};
      \node[dynnode] (c3) at (0.48, 0.20) {$3$};
      \node[dynnode] (c4) at (0, -0.43) {$4$};
      \draw[weakedge] (c1) -- (c2);
      \draw[weakedge] (c1) -- (c3);
      \draw[weakedge] (c1) -- (c4);
      \draw[weakedge, dashed] (c2) -- (c3);
      \draw[hodyn] (c2) to[bend right=22] (c4);
      \node[font=\tiny, align=center, anchor=north] at (0, -0.62) {$T_3$: higher\\$c_{24|13}$ Gumbel};
    \end{scope}
  \end{scope}
  
  \draw[gray!50, dashed] (7.75, -1.1) -- (7.75, 3.4);
  
  \end{tikzpicture}
  }
  
  \vspace{2mm}
  
  \resizebox{0.99\linewidth}{!}{%
  \begin{tikzpicture}[
    font=\small,
    node distance=5mm,
    every node/.style={font=\small},
    box/.style={draw, thick, rounded corners=2pt, minimum height=9mm, minimum width=24mm, align=center, inner sep=3pt},
    arrowstyle/.style={-{Latex[length=2mm]}, thick},
    panel/.style={font=\bfseries\small},
  ]
  
  \node[panel, anchor=west] at (0, 2.2) {(C) DVC pipeline and diagnostics};
  
  \node[box, fill=blue!8] (data)  at (0.8, 1.0) {Data \\ $\{X^{(t)}\}_{t=1}^{T}$};
  \node[box, fill=blue!8, font=\scriptsize] (pseudo)  at (4.2, 1.0) {Pseudo-obs. \\[-1pt] $\scriptstyle u_{n,i}^{(t)}=r_{n,i}^{(t)}/(N_t+1)$};
  \node[box, fill=yellow!15] (fit) at (7.6, 1.0) {Joint DVC \\ smooth / switch};
  \node[box, fill=red!10] (readout) at (11.2, 1.0) {Readouts \\ $\widehat S_{\mathrm{total}}$, $\widehat\Delta_{\mathrm{HO}}$, families};
  
  \draw[arrowstyle] (data) -- (pseudo);
  \draw[arrowstyle] (pseudo) -- (fit);
  \draw[arrowstyle] (fit) -- (readout);
  
  \node[anchor=west, font=\footnotesize] at (0.1, -0.4)
    {$\TC_t
      =
      \textstyle\sum_{(i,j)\in T_1}\MI(X_i^{(t)}, X_j^{(t)})
      +
      \textstyle\sum_{m\geq 2}\sum_{e\in T_m}\MI(X_i^{(t)}, X_j^{(t)}\mid X_D^{(t)});
      \quad
      \widehat{\Delta}_{\mathrm{HO}}
      =
      \mathrm{NLL}_{\text{1-trunc}}-\mathrm{NLL}_{\mathrm{full}}$};
  
  \end{tikzpicture}
  }
  \caption{\textbf{DVC overview.}
  \textbf{(A)}~A 4-variable C-vine with root order $(1,3,2,4)$ decomposes the copula into first-tree pairwise edges, second-tree conditional edges, and a third-tree conditional edge.
  \textbf{(B)}~DVC fixes one vine factorization across time and lets edge states evolve within it. DVC-smooth assigns each edge one family and a smooth parameter path; DVC-switch selects a temporally regularized family/parameter path. Highlighted edges illustrate shifts in predictive contribution across tree levels, not independent topology refits.
  \textbf{(C)}~Raw observations are converted to within-window pseudo-observations, fit by one joint temporal vine, and evaluated by held-out scores. The main readout is $\widehat{\Delta}_{\mathrm{HO}}=\mathrm{NLL}_{\text{1-trunc}}-\mathrm{NLL}_{\mathrm{full}}$, which equals the higher-tree total-correlation term at the population truth and is used here as a finite-sample predictive diagnostic.}
  \label{fig:overview}
  \end{figure}

  Most dynamic-dependence tools answer only part of this question.
  Time-varying correlations, dynamic conditional correlation models \citep{Engle2002}, and temporally regularized Gaussian graphical models \citep{Hallac2017} describe evolving covariance or precision structure, but remain tied to second-order, Gaussian-style dependence.
  They can track correlations that strengthen, weaken, or reorganize, but cannot directly represent changes in dependence \emph{type}: lower-tail to upper-tail dependence with little change in Kendall's $\tau$, symmetric to asymmetric dependence, or conditional interactions that emerge while pairwise correlations remain weak or stable.
  Thus, a correlation trajectory can appear stable even when the joint distribution has changed substantially.
  
  Copulas provide a natural way to go beyond correlation by separating marginal behavior from dependence structure.
  Vine copulas \citep{Bedford2002,Aas2009,Czado2019} are especially useful because they factorize a multivariate copula density into bivariate pair-copulas arranged on linked trees.
  Each edge can choose its own family, allowing one model to combine near-Gaussian, heavy-tailed, and asymmetric dependence.
  The tree structure also provides an operational order distinction: first-tree edges describe pairwise dependence, while higher-tree edges describe conditional pair-copulas after recursive conditioning.
  Vines are therefore useful not only as flexible density models, but also as tools for asking where in the dependence factorization a signal appears.
  
  Standard vine-copula analyses, however, are mostly static.
  They can reveal that dependence is non-Gaussian on average, but not when it changes, whether the changing signal is pairwise or conditional, or whether a temporal pattern is coherent across windows.
  Dynamic vine formulations exist \citep{Patton2006IER,Patton2006JAE,KreuzerCzado2019,FDCV2019,NaglerKrugerMin2021}, but they have primarily been developed as econometric density or forecasting models.
  They are less often used to diagnose the order of a changing dependence signal, and are rarely evaluated in controlled regimes that separate Gaussian pairwise changes, non-Gaussian tail changes, family switches, and conditional interactions.
  
  Here, we use dynamic vines as a temporal diagnostic rather than introduce a new static copula family.
  Given a fixed C-, D-, or R-vine factorization, DVC\dvcCodeFootnote{} couples pair-copula families and parameters across time so that all windows share one comparable dependence object.
  In experiments we use fixed-root-order C-vines.
  DVC-smooth represents gradual edge changes with smooth parameter trajectories, whereas DVC-switch represents abrupt or recurrent changes with temporally regularized family/parameter paths.
  Held-out comparisons with Gaussian dynamic baselines, windowed-vine controls, and matched 1-truncated vines then ask whether observed changes are Gaussian, non-Gaussian but pairwise, or higher-tree conditional.
  The benchmark suite includes controlled tail changes, family switches, recurrent interaction episodes, a main Allen VBN temporal analysis, and an appendix Dalgleish robustness study.

  \section{Background: Vine Copulas}
  
  For continuous marginals $F_1,\dots,F_d$, Sklar's theorem factorizes the joint density as $f(\bx)=c(\bu)\prod_{i}f_i(x_i)$ where $u_i = F_i(x_i) \in (0,1)$ and $c$ is the copula density.
  Vine copulas express $c$ as a product of bivariate terms on $d-1$ linked trees:
  \begin{equation}\label{eq:vine}
  \log c(\bu)=\sum_{m=1}^{d-1}\sum_{(i,j\mid D)\in E_m}\log c_{ij\mid D}\bigl(u_{i\mid D},\;u_{j\mid D};\;\theta_{ij\mid D}\bigr),
  \end{equation}
  where $E_m$ is the edge set in tree $m$, $D$ is the conditioning set, $\theta_{ij|D}$ are edge-specific parameters, and $h(v\mid u) = \partial C(u,v)/\partial u$ is the h-function (conditional CDF).
  Equivalently, $u_{i|D}=F_{i|D}(x_i\mid x_D)$ denotes a conditional pseudo-observation computed recursively through these h-functions.
  Three standard topologies exist: C-vines, D-vines, and R-vines.
  Each edge independently selects a bivariate copula family, enabling heterogeneous dependence types within a single joint model.
  In practice, we use rank-based pseudo-observations $u_{n,i}=\mathrm{rank}(x_{n,i})/(N+1)$ and select static pair-copula families by the Akaike information criterion (AIC) \citep{Akaike1974}.
  The simplifying assumption---that conditional pair-copulas depend only on pseudo-observations, not conditioning values---makes sequential estimation tractable; non-simplified approaches exist but at higher computational cost \citep{Bedford2002,Aas2009,Czado2019,Dissmann2013,Joe2014,Killiches2016,Nagler2024,SchellhaseSpanhel2016,Sklar1959}.
  The static construction above is the object we temporalize: DVC keeps the pair-copula factorization but changes how edge families and parameters are estimated across time.
  
  \section{Dynamic Vine Copulas}
  \label{sec:dynamic}
  
  \subsection{Temporal Model}
  
  \label{sec:problem_formulation}
  Consider observations indexed by time windows,
  $\bx^{(t)}=\{x_n^{(t)}\}_{n=1}^{N_t}\subset\R^d$ for $t=1,\dots,T$.
  At each time $t$, the dependence structure is represented by a copula density $c_t$.
  A fully general dynamic vine could allow three objects to change over time: edge parameters, edge families, and vine topology.
  Estimating all three independently in every window would be noisy and hard to compare, so DVC fixes a sequence-wide vine factorization and couples its edge states across time.
  
  For a fixed edge set $\mathcal{V}=\{E_m\}_{m=1}^{d-1}$, DVC represents
  \begin{equation}\label{eq:dynamic_copula}
  \log c_t(\bu)
  =
  \sum_{m=1}^{d-1}
  \sum_{e=(i,j\mid D)\in E_m}
  \log c_{e,t}
  \!\left(
  u_{i\mid D,t},
  u_{j\mid D,t};
  \theta_e(t)
  \right),
  \end{equation}
  where each edge state contains a copula family and parameter at time $t$.
  The framework is topology-agnostic: the same temporal parameterizations can be used with C-, D-, or R-vines once the fixed factorization is chosen.
  For stable estimation and controlled comparison, all experiments instantiate DVC with fixed-root-order C-vines.
  Windowed and regularized-windowed vines may refit local structures as controls, but the primary DVC models are sequence-wide fixed-factorization temporal vines.
  
  \label{sec:joint_dynamic}
  DVC uses two complementary edge-state parameterizations.
  \textbf{DVC-smooth} targets gradual changes in dependence strength or tail behavior.
  For edge $e=(i,j\mid D)$ with family $F_e$, it uses
  \begin{equation}
  \label{eq:smooth_param}
  \eta_e(t) \in \R^{p_{F_e}}, \qquad
  \eta_e(t) = B(t)^\top \beta_e, \qquad \theta_e(t) = g_{F_e}\!\bigl(\eta_e(t)\bigr),
  \end{equation}
  where $B(t)\in\R^q$ is a low-dimensional time basis, $\beta_e\in\R^{q\times p_{F_e}}$ are shared coefficients fit across all $t$, and $g_{F_e}$ maps unconstrained latent values to valid family parameters.
  For one-parameter families $p_{F_e}=1$; for Student-$t$ edges, DVC-smooth can either select a fixed degrees-of-freedom value from a grid or fit both Kendall-$\tau$/correlation and degrees of freedom as smooth trajectories, depending on the experiment configuration.
  For a fixed family, the trajectory is optimized with the penalized sequence objective
  \begin{equation}\label{eq:objective_joint}
  \sum_{t=1}^{T} \text{NLL}_{e,t}\!\bigl(F_e,\theta_e(t)\bigr) + \lambda_{\mathrm{smooth}} \|\Delta^2 \tau_e\|_2^2 + \lambda_{\mathrm{ridge}}\|\beta_e\|_2^2,
  \end{equation}
  where $\mathrm{NLL}_{e,t}$ is the summed training NLL for edge $e$ at time $t$, $\tau_e(t)$ is the implied Kendall-$\tau$ path, and $\Delta^2$ denotes the second finite difference on the observed time grid.
  The family $F_e$ is then selected globally by a BIC-style sequence score, $\mathrm{NLL}+ \tfrac{1}{2}k_e\log(\sum_t N_t)$, where $k_e$ is the number of edge-trajectory coefficients and any additional Student-$t$ degrees-of-freedom parameter.
  Thus DVC-smooth estimates one smooth temporal trajectory per edge rather than fitting and smoothing unrelated per-window copulas.
  
  \textbf{DVC-switch} targets abrupt family changes and recurrent interaction episodes.
  For edge $e$, let $s_{e,t}=(F_{e,t},\theta_{e,t})$ denote a candidate pair-copula state at time $t$, obtained by fitting the candidate families on the current pseudo-observations.
  The selected path is the dynamic-programming solution
  \begin{equation}\label{eq:switching_objective}
  \min_{s_{e,1:T}}
  \sum_{t=1}^T \frac{\mathrm{AIC}_{e,t}(s_{e,t})}{2N_t}
  + \lambda_{\mathrm{sw}}\sum_{t=2}^T \mathbf{1}\{F_{e,t}\neq F_{e,t-1}\}
  + \lambda_{\mathrm{drift}}\sum_{t=2}^T \mathbf{1}\{F_{e,t}=F_{e,t-1}\}\,d(\theta_{e,t},\theta_{e,t-1}).
  \end{equation}
  Here $\mathrm{AIC}=2\mathrm{NLL}+2k$, so $\mathrm{AIC}/(2N_t)=(\mathrm{NLL}+k)/N_t$ puts the local cost on a per-observation NLL-plus-parameter-penalty scale.
  DVC-switch is therefore a joint temporal state-selection model over locally fitted candidate states, not a fully joint continuous optimization of all edge parameters.
  After each tree level is fit, the selected edge copulas propagate $h$-function pseudo-observations to the next level, so both DVC-smooth and DVC-switch remain full multivariate vine models.
  A shared low-rank latent-state variant, in which multiple edges borrow temporal strength through a common state, is described in Appendix~\ref{app:latent_state}; it is reported as an ablation rather than as the main estimator.
  
  \subsection{Diagnostics and Controls}
  
  \label{sec:windowed}
  We use windowed vine estimators only as controls.
  The \emph{windowed} estimator fits an independent vine in each time window from rank-based pseudo-observations, selecting a C-vine root by
  \begin{equation}\label{eq:root_selection}
  r^*(t)=\arg\max_{r}\sum_{j\neq r}\left|\hat\tau_{rj}^{(t)}\right|,
  \end{equation}
  and selecting edge families by AIC.
  The \emph{regularized} estimator keeps the same per-window fit but adds penalties on root changes, family switches, and parameter drift between adjacent windows.
  Its objective is the windowed analogue of Eq.~\eqref{eq:switching_objective}: local AIC costs plus $\lambda_{\mathrm{root}}\mathbf{1}\{r_t\neq r_{t-1}\}$, $\lambda_{\mathrm{sw}}\mathbf{1}\{F_{e,t}\neq F_{e,t-1}\}$, and $\lambda_{\mathrm{drift}}d(\theta_{e,t},\theta_{e,t-1})$ terms.
  These controls are intentionally strong, but they produce local refit sequences rather than one jointly estimated temporal dependence object.
  Throughout the paper, only jointly fitted temporal vines are called DVC: DVC-smooth is the smooth-basis model, DVC-switch is the switching-state model, and DVC-latent is the low-rank temporal ablation.
  Win. vine denotes independent per-window refits, and Reg. win. denotes the corresponding temporally regularized windowed control.
  Table~\ref{tab:model_names} gives the same definitions in tabular form.
  
  \begin{table}[t]
  \centering
  \caption{\textbf{Model names used in all figures.}
  DVC denotes a jointly fitted temporal vine. Windowed and regularized-windowed vines are controls, not DVC models.}
  \label{tab:model_names}
  \small
  \setlength{\tabcolsep}{4pt}
  \resizebox{\linewidth}{!}{%
  \begin{tabular}{p{0.15\linewidth}p{0.18\linewidth}p{0.44\linewidth}p{0.15\linewidth}}
  \toprule
  Figure label & Role & Definition & Score readout \\
  \midrule
  DVC-smooth & primary DVC for smooth dynamics & Joint smooth-basis full vine; one sequence-wide fixed-factorization vine (C-vine in experiments) with smooth edge-parameter trajectories $B(t)^\top\beta_e$. & primary for continuous trajectories \\
  DVC-switch & primary DVC for abrupt dynamics & Joint switching-state full vine; one sequence-wide fixed-factorization vine (C-vine in experiments) with edge family/parameter paths selected over time. & primary for family switches and episodes \\
  Win. vine & control & Independent static full vine fit separately in each time window; no joint temporal state or shared trajectory. & only when explicitly labeled ``Win. vine'' \\
  Reg. win. & control & Windowed full-vine sequence with penalties on adjacent root/family/parameter changes. & only when explicitly labeled ``Reg. win.'' \\
  \bottomrule
  \end{tabular}
  }
  \end{table}
  
  \label{sec:change_detection}
  All methods are evaluated by held-out copula negative log-likelihood (NLL).
  For any baseline $b$, we use
  \begin{equation}\label{eq:nll_gap}
  \Delta(b,t)=\mathrm{NLL}_b(t)-\mathrm{NLL}_{\mathrm{DVC}}(t),
  \end{equation}
  so positive values indicate that DVC gives higher held-out likelihood.
  Unless otherwise stated, $\mathrm{NLL}_{\mathrm{DVC}}$ refers to the primary temporal model for the regime: DVC-smooth for continuous trajectories and DVC-switch for abrupt switches or episodes.
  
  The central diagnostic compares a full vine with its matched 1-truncated counterpart.
  The two models share the same first-tree pairwise edges; the 1-truncated model removes all higher-tree levels.
  The score difference asks whether, after allowing flexible non-Gaussian pairwise copulas, conditional vine levels still improve prediction.
  For a fitted model $M$, we report
  \begin{equation}\label{eq:tc_decomp}
  \begin{aligned}
  \widehat S_{\mathrm{total}}^M(t)
  &=
  -\mathrm{NLL}(M_{\mathrm{full}},t),\\
  \widehat S_{\mathrm{pair}}^M(t)
  &=
  -\mathrm{NLL}(M_{\text{1-trunc}},t),\\
  \widehat \Delta_{\mathrm{HO}}^M(t)
  &=
  \mathrm{NLL}(M_{\text{1-trunc}},t)-\mathrm{NLL}(M_{\text{full}},t).
  \end{aligned}
  \end{equation}
  A positive $\widehat\Delta_{\mathrm{HO}}^M(t)$ means that higher-tree vine terms improve held-out prediction beyond the matched first-tree truncation.
  At population truth, under a correct fixed vine and the simplifying assumption, this contrast equals the higher-tree component of a vine total-correlation decomposition.
  Because this is a fixed-vine, simplified-vine diagnostic, a positive $\widehat\Delta_{\mathrm{HO}}^M(t)$ should be read as higher-tree predictive content for the chosen factorization, not as topology-free proof of irreducible multiway interaction.
  In the fixed C-vine instantiation, dense pairwise structure outside the selected first-tree star can contribute to higher trees, and the pairwise-matched XOR stress test shows that some higher-order constructions can be missed.
  We therefore interpret small negative held-out estimates as lack of reliable higher-tree evidence rather than negative dependence, and we use root-order sensitivity checks or structure-agnostic readouts when feasible.
  In figures, unqualified pair and higher-tree score labels always refer to these held-out estimates for the primary DVC model in that panel; windowed-control quantities are plotted only when explicitly labeled ``Win. vine''.
  The three recurring comparisons answer different questions: Gaussian gaps test whether second-order dynamics suffice, the full-vs-1-truncated gap tests whether higher-tree vine terms matter for the chosen factorization, and windowed-vine gaps test what is lost or gained by one temporal model relative to local refits.

  Theoretical support is intentionally targeted rather than expansive.
  Appendix~\ref{app:proofs} states the total-correlation decomposition, identifiability of the joint basis estimator under fixed structure and family, and an oracle consistency result for the same fixed-structure setting.
  We compare against static Gaussian copulas, Graphical Lasso \citep{Friedman2008GraphicalLasso}, a time-varying graphical lasso (TVGL)-style Gaussian precision sequence, a Gaussian state-space model (Gaussian SSM) baseline on Fisher-$z$ edge dynamics, and selected learned baselines---normalizing-flow copulas (NF-copulas) and Mutual Information Neural Estimation (MINE)---where they sharpen interpretation.
  This separation is essential: a gap over Gaussian baselines says that dynamic second-order structure is insufficient, whereas a gap over the matched 1-truncated vine is evidence that higher-tree terms add predictive information for the chosen vine.
  Dynamic vine copulas and time-varying pair-copula parameters have been studied previously \citep{Patton2006IER,Patton2006JAE,FDCV2019,KreuzerCzado2019,NaglerKrugerMin2021}; DVC uses this modeling family for a time-resolved held-out order diagnostic rather than only density estimation or forecasting.
  Similarly, truncated-vine ideas are classical in static vine modeling \citep{Brechmann2012Truncated,BrechmannJoe2015Truncation}; our use of a matched full-vs-1-truncated contrast is a predictive temporal readout attached to one sequence-wide dynamic model.
  \ifdvcneurips
\else
  Static neural copula estimators provide useful complementary context: classifier-based ratio copulas \citep{Huk2025Classifier} learn the copula density as a joint-versus-independent density ratio, diffusion- and flow-based copulas \citep{HukDamoulas2026} construct global copulas through marginal-preserving noising processes, and amortized vine copulas \citep{Safaai2026Amortized} replace repeated static pair-copula fitting with a reusable neural edge estimator.
  Related neural population work has also used task- or time-indexed analyses of dependence structure \citep{Safaai2025NatNeurosci}, and copula models have been used directly on neural responses \citep{Berkes2008,Onken2016,Safaai2018PRE,Mitskopoulos2022,Mitskopoulos2023,Kudryashova2022}.
  \fi
  Order-sensitive readouts complement multivariate-mutual-information decompositions of higher-order interdependencies \citep{Rosas2019} and connect to recent vine-copula generative \citep{Tagasovska2019,Tagasovska2023} and differentiable-vine implementations \citep{torchvinecopulib2025}.
  Here we instead fit a sequence-wide dynamic copula state path: time is not only a covariate inside one copula surface, but the index of the jointly estimated family and parameter trajectory.
  In contrast to static learned copulas, DVC focuses on temporal variation, edge-parameter trajectories, and order-sensitive interaction diagnostics rather than static global copula density estimation.
  Detailed baseline equations, algorithmic specifics, and extended related-work discussion are deferred to Appendix~\ref{app:baseline_details}.
  
  \section{Experiments}
  \label{sec:experiments}
  
  The experiments are designed to test DVC as an order-sensitive diagnostic rather than only as a density estimator.
  They follow three questions.
  First, can a temporal copula detect dependence changes that leave Gaussian correlations nearly unchanged?
  Second, when a change is detected, is it explained by flexible non-Gaussian first-tree pairwise terms, or does it require higher-tree conditional terms?
  Third, does sequence-wide temporal coupling recover these signals as one coherent time-indexed model rather than as independent local refits?
  
  The controlled suite isolates four mechanisms.
  The first two are non-Gaussian but primarily pairwise: a piecewise Student-$t$ degrees-of-freedom change at fixed correlation, and an abrupt Clayton-to-Gumbel family switch at fixed Kendall's $\tau$.
  These test whether DVC detects changes in dependence type that Gaussian dynamic baselines cannot represent, while the matched 1-truncated vine should remain competitive if the signal is truly pairwise.
  The third mechanism introduces recurrent agent episodes with pairwise, higher-tree, and mixed interaction epochs, testing whether the full-vs-1-truncated contrast assigns the correct interaction order over time.
  The fourth is a $d=10$ four-phase showcase combining independence, Gaussian pairwise dependence, multiplicative-triplet structure, and lower-tail dependence in one sequence.
  We also include a pairwise-matched XOR stress test as a limitation check: pairwise summaries are nearly unchanged by construction, and the current simplified C-vine estimator should not be assumed to detect every possible higher-order interaction.
  
  \begin{figure}[!t]
  \centering
  \includegraphics[width=\linewidth]{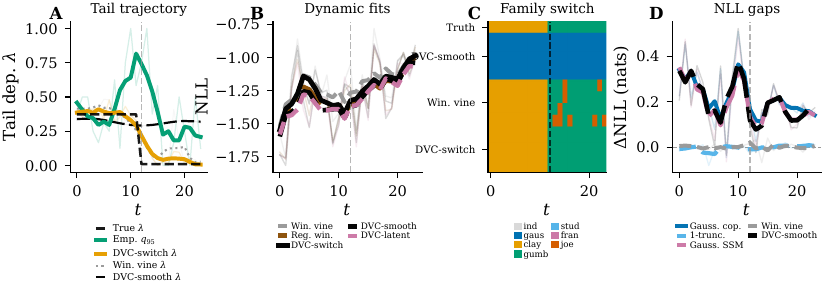}
  \caption{\textbf{Tail and family-switch dynamics.} \textbf{(A)}~Piecewise Student-$t$ degrees-of-freedom at fixed correlation: analytic truth, empirical tail summary, and fitted tail-dependence trajectories. \textbf{(B)}~Held-out copula NLL (lower is better) for temporal variants on the same tail sequence. \textbf{(C)}~Clayton-to-Gumbel switch at fixed Kendall's $\tau$: DVC-switch represents the family change as one temporally regularized state path, while the Win. vine tracks it through independent local refits. \textbf{(D)}~Baseline gaps show that the switch is non-Gaussian but primarily first-tree pairwise, as the matched 1-truncated vine remains competitive while Gaussian baselines lose likelihood.}
  \label{fig:sim_tail}
  \end{figure}
  \ifdvcneurips\else\clearpage\fi
  
  Baselines are chosen to make failures interpretable rather than easy.
  Gaussian copulas, Graphical Lasso, TVGL-style precision sequences, and Gaussian state-space models test whether dynamic second-order structure is sufficient.
  The matched 1-truncated vine tests whether flexible non-Gaussian first-tree dependence is sufficient.
  The independent Win. vine and Reg. win. controls test whether a joint temporal fit merely reproduces local static refits.
  Selected learned baselines, Real-NVP/NF-copula density models \citep{Dinh2017RealNVP,Laszkiewicz2021,Rezende2015} and MINE \citep{Belghazi2018}, are used only in the showcase where they sharpen interpretation.
  
  All models are evaluated by held-out copula NLL gaps using Eq.~\eqref{eq:nll_gap}, with positive values favoring the named DVC estimator.
  The primary estimator is fixed before evaluation by the expected temporal regime: DVC-smooth for continuous trajectories and DVC-switch for abrupt family changes or recurrent episodes.
  For order assignment, every full-vine model is compared with its own matched 1-truncation using Eq.~\eqref{eq:tc_decomp}; unless explicitly labeled otherwise, pair and higher-tree curves refer to the primary joint DVC readouts.
  Synthetic $\pm$ values summarize variability over time windows for the fixed benchmark seed, not independent regenerated datasets.
  Allen uncertainty is computed over session/split fits.
  Additional benchmark definitions, split details, hyperparameters, and full tables are given in Appendices~\ref{app:method_details}--\ref{app:full_bench}.
  
  \section{Results}
  \label{sec:results}
  
  We first test non-Gaussian regimes whose dependence changes are intentionally pairwise.
  These cases serve as negative controls for the higher-tree diagnostic: DVC should outperform Gaussian dynamic baselines, but the matched 1-truncated vine should remain competitive if first-tree pairwise copulas are sufficient.
  In the piecewise Student-$t$ degrees-of-freedom benchmark (Figure~\ref{fig:sim_tail}A--B), the linear correlation is fixed while tail dependence changes.
  DVC-smooth captures the broad tail-state shift and improves held-out copula NLL relative to the Gaussian state-space baseline ($+0.153$ nats) and the independent Win. vine control ($+0.094\pm0.065$ nats).
  In the Clayton-to-Gumbel switch at fixed Kendall's $\tau$ (Figure~\ref{fig:sim_tail}C--D), DVC-switch recovers the change in copula family and improves over Gaussian SSM ($+0.195\pm0.127$ nats).
  At the same time, its near tie with the matched 1-truncated vine ($-0.002\pm0.024$ nats) gives the intended interpretation: the switch is a non-Gaussian pairwise dependence change, not higher-tree evidence.
  
  The pairwise-matched XOR stress test gives the complementary limitation.
  Although the mean pairwise-correlation shift is only $0.0027$, the simplified C-vine estimator also gives near-null higher-tree evidence (AUROC $0.50$).
  We therefore treat this benchmark as a failure mode rather than a positive result: the proposed diagnostic is useful for vine-representable conditional structure, but should not be assumed to detect every possible higher-order interaction.
  
  \begin{figure}[!t]
  \centering
  \ifdvcneurips
\else
  \includegraphics[width=\linewidth]{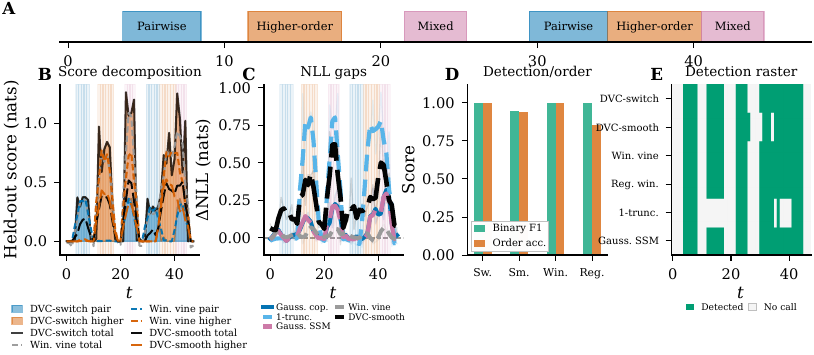}
  \fi
  \caption{\textbf{Agent interaction episodes.} \textbf{(A)}~Ground-truth recurrent schedule. \textbf{(B)}~DVC-switch held-out score decomposition, with matched Win. vine and DVC-smooth controls shown only where explicitly labeled. \textbf{(C)}~NLL gaps against Gaussian, 1-truncated, Win. vine, and smooth-DVC comparators. \textbf{(D--E)}~Detection and order assignment over time windows: the main advantage is not binary detection, but separating pairwise from higher-tree episodes.}
  \label{fig:sim_episodes}
  \end{figure}
  
  After these pairwise controls, the recurrent agent benchmark tests whether DVC can assign detected dependence to the correct vine order.
  Several methods identify parts of the active schedule because some episodes contain strong pairwise correlation.
  The distinctive question is whether the model assigns the active dependence to the correct order.
  DVC-switch improves over Gaussian SSM ($+0.070\pm0.118$ nats) and over its matched 1-truncated vine ($+0.286\pm0.378$ nats), but the key result is the temporal selectivity of $\widehat\Delta_{\mathrm{HO}}$.
  The higher-tree gain remains near zero in pairwise-only epochs ($-0.018$ nats), rises in higher-tree epochs ($+0.683$ nats), and remains high in mixed epochs ($+0.744$ nats).
  Thus the full-vs-1-truncated contrast distinguishes whether a detected interaction is explained by first-tree pairwise terms or by higher-tree terms in the chosen vine, a distinction unavailable to Gaussian baselines, pairwise MI readouts, or the 1-truncated vine alone.
  
\begin{figure}[!htb]
  \centering
  \ifdvcneurips
\else
  \includegraphics[width=\linewidth]{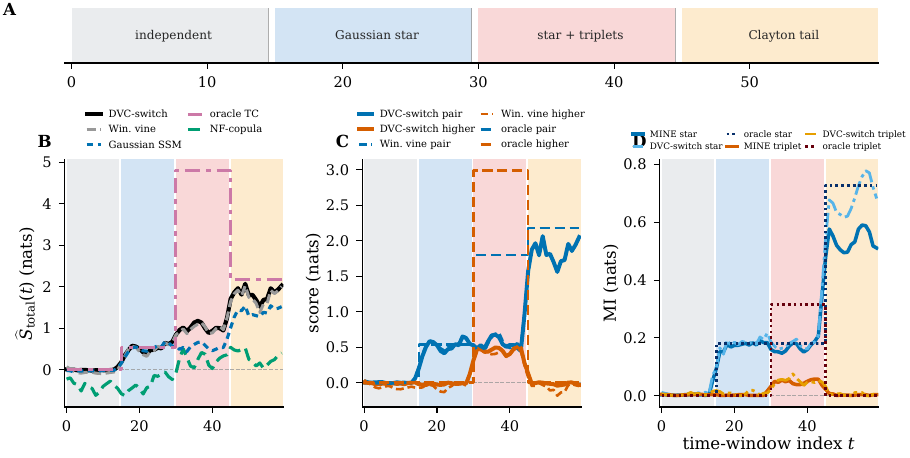}
  \caption{\textbf{Four-phase detection showcase} ($d=10$, $T=60$), showing DVC-switch together with the independent Win. vine control. Mean trajectories over available seeds; shaded bands show seed-to-seed standard deviation when multiple seeds are available. DVC and Win. vine curves are held-out score estimates; dashed oracle curves are population information values. This figure emphasizes relative detection and order decomposition; the oracle overlay exposes absolute underestimation in the near-deterministic triplet phase. \textbf{(A)}~Ground-truth phase timeline. \textbf{(B)}~Held-out total score estimate $\widehat S_{\mathrm{total}}(t)$ from DVC-switch, Win. vine, NF-copula, and Gaussian state-space against oracle TC. \textbf{(C)}~DVC-switch score decomposition $\widehat S_{\mathrm{total}}=\widehat S_{\mathrm{pair}}+\widehat\Delta_{\mathrm{HO}}$ with matched Win. vine pair and higher-tree controls: the pairwise star phase is absorbed by first-tree edges, the multiplicative triplet phase lifts $\widehat\Delta_{\mathrm{HO}}$, and the Clayton tail phase returns it toward zero. \textbf{(D)}~Pairwise MI for a star pair $(X_0, X_1)$ and a triplet-internal pair $(X_5,X_6)$: MINE and sample-based DVC-switch MI are compared with oracle pairwise MI.}
  \fi
  \label{fig:showcase}
  \end{figure}

  The four-phase $d=10$ showcase (Figure~\ref{fig:showcase}) then combines Gaussian, pairwise non-Gaussian, and higher-tree regimes in a single sequence: independence, Gaussian pairwise dependence, multiplicative-triplet structure, and lower-tail dependence.
  DVC-switch behaves as desired across these regimes.
  It approximately matches Gaussian SSM in the Gaussian star phase (DVC-switch minus SSM $\approx -0.01$ nats), separates from Gaussian SSM in the non-Gaussian triplet and Clayton phases (both about $+0.46$ nats), and localizes higher-tree evidence to the multiplicative-triplet phase.
  Specifically, $\widehat\Delta_{\mathrm{HO}}$ is near zero in the pairwise and lower-tail phases ($-0.020$ and $-0.017$ nats), but positive in the multiplicative-triplet phase ($+0.458$ nats).
  The oracle TC is deliberately large for the near-deterministic triplet phase, so this figure should be read as a detection and decomposition test rather than exact recovery of all oracle information.
  
  \ifdvcneurips
\else
  The computational motivation for joint DVC is not that the evaluated implementation must beat the Win. vine in wall-clock time on every small CPU benchmark; the Win. vine solves many simple independent fits, whereas the joint estimator evaluated here uses generic unvectorized trajectory optimization.
  The key scaling difference is in the number of fitted dependence objects.
  A $d$-dimensional C-vine has $d(d-1)/2$ pair-copula edges.
  The Win. vine refits all edges independently at each of $T$ time points, for $T\,d(d-1)/2$ edge fits, whereas smooth joint DVC fits $d(d-1)/2$ temporal edge trajectories.
  Appendix Table~\ref{tab:runtime} quantifies this directly: for $d=8,T=24$, the Win. vine fits $672$ edge objects, while smooth joint DVC fits $28$ trajectories, a $24\times$ compression.
  The measured CPU times are total sequential times for the entire 24-window sequence, not single-window times: Win. vine takes $21.34$s total, or $0.889$s per fitted window on average, while joint DVC takes $26.49$s for one complete temporal fit over all windows.
  Thus windowed fitting is faster in this CPU benchmark because each local static fit is cheap, even though its cost and number of fitted models grow linearly with $T$.
  Appendix~\ref{app:runtime} gives the formal scaling and crossover condition: joint DVC does not remove the need to evaluate all $TN$ samples, but it can win in wall-clock time when repeated per-window optimization overhead dominates or when the joint trajectory optimizer is vectorized/uses analytic gradients.
  We therefore do not claim an implemented wall-clock speedup yet; we claim a substantially more compact temporal parameterization and a clear optimization target.
  \fi
  
\begin{figure}[!t]
  \centering
  \includegraphics[width=\linewidth]{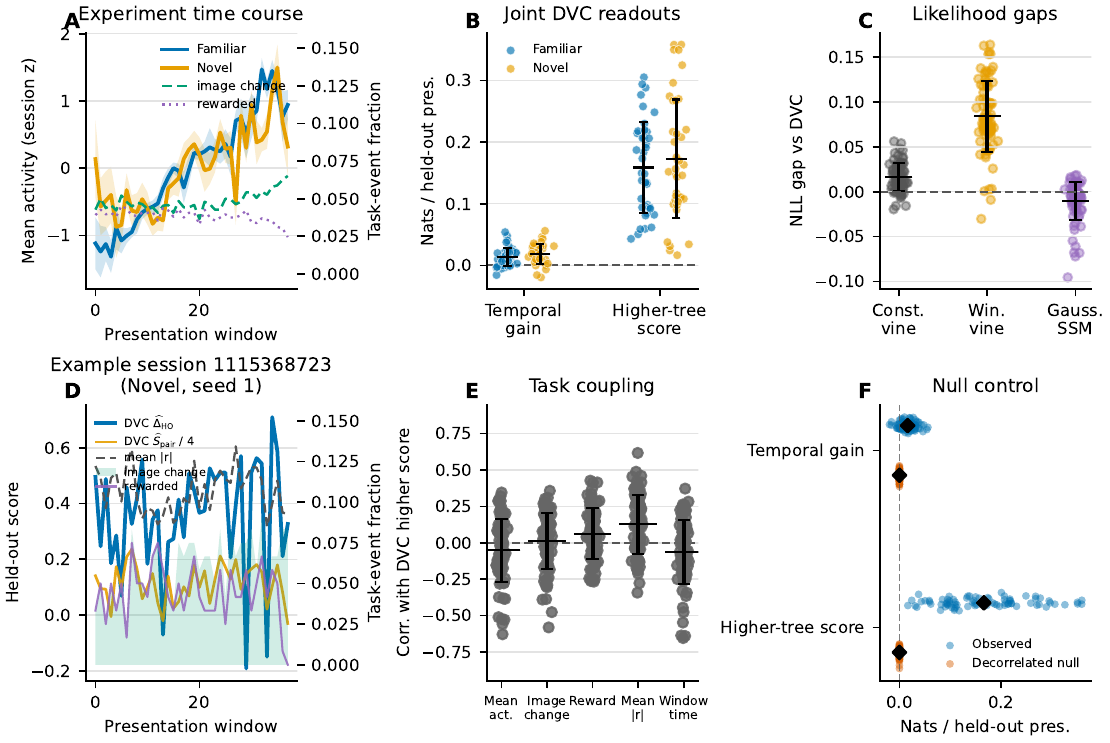}
  \ifdvcneurips
\else
  \caption{\textbf{Allen Visual Behavior Neuropixels (VBN) joint-DVC temporal validation.}
  Dots denote session-seed fits from five random held-out splits of the same presentation-order windows.
  \textbf{(A)}~Experiment time course: windowed mean post-stimulus activity for familiar and novel sessions (session-wise $z$ score), with image-change and rewarded-event fractions on the right axis.
  \textbf{(B)}~Joint-DVC temporal NLL gain and higher-tree held-out score $\widehat\Delta_{\mathrm{HO}}=\mathrm{NLL}_{\text{1-trunc}}-\mathrm{NLL}_{\mathrm{full}}$, in nats per held-out presentation.
  \textbf{(C)}~NLL gaps relative to joint DVC for the constant vine, independent Win. vine, and Gaussian SSM.
  \textbf{(D)}~Representative session timeline showing DVC pair and higher-tree score estimates, mean absolute correlation, and task-event fractions; $\widehat S_{\mathrm{pair}}$ is divided by 4 for visualization.
  \textbf{(E)}~Correlations between DVC higher-tree score and activity/task/correlation covariates; coupling to image changes and rewards is modest rather than dominant.
  \textbf{(F)}~Decorrelated-null control for the two DVC readouts in panel B; temporal gain and higher-tree score collapse when marginal responses and time sampling are preserved but simultaneous cross-area interactions are removed.}
  \fi
  \label{fig:allen_joint_dvc}
  \end{figure}

  Across the synthetic suite, the pattern is consistent: DVC gains are clearest when the data contain non-Gaussian dependence, while $\widehat\Delta_{\mathrm{HO}}$ is most informative when the ground truth contains conditional higher-tree structure.
  We finally apply the same diagnostic to Allen VBN to test whether this behavior yields a stable neural-data signal.
  \ifdvcneurips
\else
  Allen Visual Behavior Neuropixels (VBN) provides a direct temporal neural test (Figure~\ref{fig:allen_joint_dvc}) \citep{AllenVBNData}.
  We fit joint DVC across 38 presentation-order windows in each of 16 sessions from 8 mice, using five visual-area population responses per session.
  Each observation is one natural-image presentation represented by the log-transformed mean spike count in each selected area during the $0$--$250$ ms post-stimulus window; the DVC time index is therefore slow presentation order, not within-trial spike time.
  In a five-seed random held-out split robustness run, the temporal model achieves a modest positive gap relative to a constant sequence-wide vine ($+0.016 \pm 0.016$ nats per held-out presentation), with positive session-seed mean in $68/80$ fits.
  It also achieves a positive gap relative to the independent Win. vine control ($+0.084 \pm 0.040$ nats; $78/80$ positive), although this comparison is partly an overfitting check because the Win. vine refits each window independently.
  The Gaussian SSM remains slightly better in held-out likelihood on this neural dataset ($-0.010 \pm 0.022$ nats for SSM-minus-joint; DVC positive in $23/80$ fits), so the Allen result should not be read as a likelihood win over all Gaussian temporal baselines.
  The DVC decomposition separates this signal into first-tree pairwise and higher-tree dependence:
  $\widehat S_{\mathrm{total}}=1.020 \pm 0.702$ nats,
  $\widehat S_{\mathrm{pair}}=0.854 \pm 0.672$ nats, and
  $\widehat\Delta_{\mathrm{HO}}=0.166 \pm 0.086$ nats per held-out presentation, so higher-tree terms account for $0.21 \pm 0.12$ of the held-out copula-TC score.
  The higher-tree component is positive in every random session-seed fit ($80/80$).
  This is not simply a linear session drift or pairwise-correlation artifact: session-level correlations between $\widehat\Delta_{\mathrm{HO}}(t)$ and mean activity, image-change fraction, rewarded fraction, mean absolute pairwise correlation, and presentation time are modest on average ($-0.051$, $0.013$, $0.063$, $0.128$, and $-0.064$, respectively).
  A larger-window robustness run with 240-presentation windows gives the same qualitative result: positive temporal gain in $48/48$ session-seed fits and positive higher-tree score in $48/48$.
  As a bias-floor control, we independently permuted each visual-area response within every train/test split and window, preserving marginal distributions and temporal sampling while destroying simultaneous cross-area interactions.
  Under this decorrelated null, joint-DVC temporal gain and both pairwise and higher-tree score estimates collapse to the independence floor ($0.000$ nats; higher-tree positive in $0/80$ fits).
  We therefore interpret Allen as a modest but null-calibrated time-indexed dependence result: joint DVC detects a higher-tree component not present after simultaneous cross-area dependence is destroyed, while the Gaussian SSM comparison indicates that likelihood alone does not require a non-Gaussian temporal model for this representation.
  The Dalgleish photostimulation analysis is kept in Appendix~\ref{app:neural_data} because it is a useful condition-level copula check, but its strength depends on the stimulation-aligned response window and it is not a fitted temporal-DVC trajectory.
  \fi
  
  \ifdvcneurips
  \else
  We keep the full Dalgleish figure, preprocessing-sensitivity discussion, and the earlier windowed Allen cohort check in Appendix~\ref{app:neural_data}, because these neural analyses are useful supporting checks but should not be read as causal circuit claims.
  \fi
  
  \section{Discussion}
  
  \ifdvcneurips
\else
  DVC is best understood as an interpretable diagnostic for time-varying dependence beyond correlation, not as a method that should always maximize held-out likelihood.
  This distinction matters because a non-Gaussian temporal change need not be higher-tree evidence.
  It helps when dependence changes in ways that time-varying Gaussian correlations cannot capture: Student-$t$ degrees-of-freedom shifts at stable correlation, abrupt family switches, and recurrent agent episodes with on/off pairwise and higher-tree interactions.
  The experiments instantiate DVC with one fixed C-vine order and rely on the simplifying assumption, so they estimate interpretable held-out score contrasts rather than a fully non-simplified dynamic copula.
  A positive higher-tree score is therefore specific to the chosen vine factorization: for a C-vine, the 1-truncated comparator is a selected first-tree star, so dense pairwise structure outside that star can also appear in higher trees.
  The pairwise-matched XOR stress test is an explicit limitation: in that construction, the current simplified-vine estimator returns near-null evidence despite the hidden triplet structure.
  In applications, we recommend checking sensitivity to several root orders or fixed C-/D-/R-vine structures, and using structure-agnostic density or information estimators as complementary stress tests when feasible.
  DVC-smooth and DVC-latent target continuous trajectories, DVC-switch targets discrete family changes, and the windowed/regularized-windowed controls test whether gains are merely local refitting.
  When dependencies are approximately Gaussian and evolve smoothly, well-regularized Gaussian dynamic baselines can be competitive at lower complexity; hence the value of full vines is not universal NLL dominance, but the extra diagnostic information from edge-family changes and $\widehat\Delta_{\mathrm{HO}}$.
  Likewise, an independent windowed vine is a high-flexibility control and can match or slightly exceed a joint dynamic fit in held-out likelihood; the advantage of joint DVC is that it estimates one coherent temporal dependence object instead of independent local refits.
  Appendix~\ref{app:runtime} quantifies this distinction: smooth joint DVC uses one temporal edge trajectory per vine edge rather than $T$ separate edge fits, although the evaluated Python implementation does not yet convert this model compression into wall-clock speedups on small CPU benchmarks.
  The neural analyses are supportive rather than mechanistic: Dalgleish is response-window sensitive, and Allen demonstrates time-indexed population dependence but does not by itself identify causal circuit motifs.
  For neural data, $\widehat\Delta_{\mathrm{HO}}$ connects the pairwise-vs-higher-order population-coding debate \citep{Schneidman2006,Ganmor2011,Panzeri2022} to a time-resolved, held-out, null-calibrated readout.
  More generally, DVC can improve dependence reporting, but a positive held-out likelihood gap or higher-tree readout does not imply causality, mechanism, or actionable clinical, policy, trading, or surveillance decisions without domain-specific validation. Experiments use synthetic and public animal-neuroscience data, with no new human-subject collection.
  \fi
  
  \clearpage
  \bibliographystyle{plainnat}
  \bibliography{dvc}

  \appendix
  
  \section{Theoretical Statements and Proofs}
  \label{app:proofs}
  
  For completeness, we collect the formal statements referenced in the main text before giving their proofs.
  
  \begin{proposition}[Population total-correlation decomposition]\label{prop:tc_decomp}
  Suppose the true copula density $c_t$ admits an exact pair-copula construction on a fixed vine structure with first tree $T_1$ and higher trees $T_{\geq 2}$.
  Under the simplifying assumption, the conditional pair-copulas do not depend on the conditioning values beyond the propagated pseudo-observations.
  Let $c_t^{(1)}$ denote the corresponding 1-truncated copula density containing only the first-tree pair-copulas.
  Then
  \[
  \TC_t(X^{(t)}) = \TC_t^{\mathrm{pair}} + \TC_t^{\mathrm{higher}},
  \]
  where
  \[
  \begin{aligned}
  \TC_t^{\mathrm{pair}}
  &= \sum_{(i,j)\in T_1}\MI(X_i^{(t)},X_j^{(t)})
     = \E[\log c_t^{(1)}(U^{(t)})],\\
  \TC_t^{\mathrm{higher}}
  &= \sum_{m\geq 2}\sum_{(i,j\mid D)\in T_m}
     \MI(X_i^{(t)},X_j^{(t)}\mid X_D^{(t)})
   = \E\!\left[\log\frac{c_t(U^{(t)})}{c_t^{(1)}(U^{(t)})}\right].
  \end{aligned}
  \]
  Both population terms are nonnegative, and $\TC_t^{\mathrm{higher}} = 0$ iff all higher-tree conditional pair-copulas are independence copulas almost surely, equivalently $c_t=c_t^{(1)}$ for this vine structure.
  \end{proposition}
  
\begin{proposition}[Identifiability of the joint basis estimator]\label{prop:identifiability}
Fix a vine structure $\mathcal{V}$, an edge-family assignment $\{F_e\}_{e\in E}$, and a time basis $B(t)$ evaluated on a grid $t \in \{1,\dots,T\}$.
If the basis design matrix has full column rank, each family link $g_{F_e}$ is injective on its latent domain, and the per-time pair-copula likelihood is identifiable at the population distribution, then the basis coefficients $\beta_e$ in $\theta_e(t)=g_{F_e}(B(t)^\top\beta_e)$ are identifiable for each edge, modulo standard family symmetries.
  \end{proposition}
  
\begin{proposition}[Oracle consistency at fixed structure and family]\label{prop:consistency}
Assume the conditions of Proposition~\ref{prop:identifiability}, with fixed $T$, fixed basis dimension $q$, true vine structure and family assignment known, and pseudo-observations treated as oracle copula samples or obtained from continuous marginals with uniformly consistent empirical CDFs.
Let $N=\sum_t N_t$.
Assume the simplifying assumption, independent observations within windows and independent or sufficiently mixing time-block contributions, twice continuously differentiable per-edge log-likelihood with bounded score and Hessian in a neighborhood of the truth, positive-definite sample-size-weighted per-edge information, controlled lower-level $h$-function propagation error under sequential estimation, $\min_t N_t \to \infty$, and a smoothness penalty satisfying $\lambda_N \to 0$ with $\sqrt{N}\lambda_N \to 0$.
  Family and structure selection are excluded from this oracle statement.
  Then the penalized joint estimator converges in probability to the true coefficient vector at the parametric rate $O_p(N^{-1/2})$, and the induced edge trajectories are pointwise consistent on the observed time grid.
  \end{proposition}

  \begin{remark}[Theory--practice scope]\label{rem:scope}
  Proposition~\ref{prop:consistency} is an oracle statement: vine structure and per-edge family are assumed fixed and correctly specified.
  The estimators in Section~\ref{sec:dynamic} and the experiments in Section~\ref{sec:results} additionally include sequence-wise family selection from a candidate set and training-set root selection for the C-vine.
  Those selection steps are not covered by the oracle result; instead, they are validated empirically by the controlled benchmark suite (tail and family-switch dynamics, multiplicative triplet, four-phase showcase, pairwise-matched XOR stress test) and by the decorrelated-null calibration on Allen.
  We therefore claim consistency only at oracle structure and family, and predictive validity under selection only through these empirical checks.
  \end{remark}

  \paragraph{Proof of Proposition~\ref{prop:tc_decomp}.}
  For a continuous random vector $X^{(t)}$ with absolutely continuous marginals, total correlation equals the expected copula log-density (equivalently, negative copula entropy), an identity noted explicitly by \citet{DavyDoucet2003} and later developed as copula entropy by \citet{MaSun2011}: $\TC_t(X^{(t)}) = \E[\log c_t(U^{(t)})]$.
Substituting the vine factorization Eq.~\eqref{eq:vine} (time-indexed) gives
  \begin{equation*}
  \TC_t = \E\Bigl[\sum_{m=1}^{d-1}\sum_{(i,j\mid D)\in T_m}\log c_{ij\mid D}^{(t)}\bigl(u_{i\mid D,t}, u_{j\mid D,t}\bigr)\Bigr].
  \end{equation*}
  Splitting the outer sum at $m=1$, the first-tree sum collects bivariate copula log-densities, each equal to a pairwise mutual information by the same identity: $\E[\log c_e(u_i,u_j)] = \MI(X_i^{(t)}, X_j^{(t)})$.
  For $m \geq 2$, under the simplifying assumption \citep{Czado2019,Joe2014}, the conditional pair-copula $c_{ij\mid D}$ depends only on the recursively propagated pseudo-observations and not on the conditioning values; hence $\E[\log c_{ij\mid D}^{(t)}(\cdot)] = \MI(X_i^{(t)}, X_j^{(t)} \mid X_D^{(t)})$.
  Both pieces are nonnegative as (conditional) mutual informations.
  $\TC_t^{\mathrm{higher}} = 0$ iff every higher-tree conditional pair-copula is the independence copula almost surely, iff $c_t \equiv c_t^{(1)}$ for the fixed vine structure. \qed
  
  \paragraph{Proof of Proposition~\ref{prop:identifiability}.}
  We argue level-by-level over $m = 1, \dots, d-1$.
  
  \emph{First tree ($m=1$).}
  Let $\Phi\in\R^{T\times q}$ be the design matrix with row $B(t)^\top$.
  For each edge $e \in T_1$, population identifiability of the per-time pair-copula likelihood gives a unique parameter $\theta_e^{(t)}$ at each $t$.
  Condition (ii) then gives a unique $\eta_e^{(t)} = g_{F_e}^{-1}(\theta_e^{(t)})$.
  Stacking $\eta_e = (\eta_e^{(1)}, \dots, \eta_e^{(T)})^\top \in \R^T$, the scalar-parameter case has parameterization $\eta_e = \Phi \beta_e$ and unique solution $\beta_e = (\Phi^\top \Phi)^{-1}\Phi^\top \eta_e$ by condition (i) (full column rank).
  For vector-valued families, stack the latent trajectories into $H_e\in\R^{T\times p_{F_e}}$.
  The parameterization $H_e=\Phi\beta_e$ is identifiable column-wise; equivalently, $\mathrm{vec}(H_e)=(I_{p_{F_e}}\otimes\Phi)\mathrm{vec}(\beta_e)$, and $I_{p_{F_e}}\otimes\Phi$ has full column rank because $\Phi$ has full column rank.
  
  \emph{Level $m \geq 2$.}
  Once levels $1, \dots, m{-}1$ are identified, the pseudo-observations propagated to level $m$ via the $h$-functions are deterministic functions of the data and the lower-level parameters.
  The same per-edge argument as in the first tree applies on these propagated pseudo-observations.
  
  By induction, all $\beta_e$ are identifiable. \qed
  
  \paragraph{Proof of Proposition~\ref{prop:consistency}.}
  The estimator is sequential, so consider one edge $e$ at a fixed tree level, conditional on the lower-level pseudo-observations available for that level.
  Its per-observation objective is
  \[
  \ell_{e,N}(\beta_e)=
  N^{-1}\sum_{t=1}^{T}\sum_{n=1}^{N_t}
  \log c_{e,t}(u_{n,i\mid D,t},u_{n,j\mid D,t};\theta_e(t;\beta_e)).
  \]
  The population Hessian is the sample-size-weighted information
  $I_e^\star=-\sum_t (N_t/N)\E_t[\nabla^2\ell_{e,t}(\beta_e^\star)]$, which is nonsingular by assumption.
  Standard finite-dimensional M-estimation theory \citep{vanderVaart2000} gives a root-$N$ edge estimator under oracle copula samples.
  With empirical CDF pseudo-observations, and with estimated lower-level $h$-function inputs, the influence function and asymptotic variance can change; the stated uniform CDF consistency and propagation-error assumptions are exactly what preserve the $O_p(N^{-1/2})$ rate claimed here.
  Because $\sqrt{N}\lambda_N\to0$, the smoothness/ridge penalty is first-order negligible for this rate.
  Applying the same argument level-by-level propagates the parametric rate up the vine.
  Continuity of $g_{F_e}$ then gives consistency of the induced edge trajectories at each grid point. \qed
  
  \section{Additional Method Details}
  \label{app:method_details}
  
  \subsection{Temporal Modeling Variants}
  \label{app:temporal_variants}
  
  Table~\ref{tab:temporal_variants} summarizes the temporal models used in the experiments.
  The DVC rows are jointly parameterized temporal vines, implemented here as fixed-root-order C-vines: DVC-smooth for continuous trajectories and DVC-switch for abrupt family/on-off changes.
  The windowed and regularized rows are controls and are not referred to as DVC.
  
  \begin{table}[h]
  \centering
  \caption{Temporal modeling variants. DVC denotes the jointly fitted temporal full-vine estimators, instantiated here with fixed-root-order C-vines; the windowed and regularized vines are controls, not DVC variants.}
  \label{tab:temporal_variants}
  \small
  \resizebox{\linewidth}{!}{%
  \begin{tabular}{p{0.20\linewidth}p{0.18\linewidth}p{0.24\linewidth}p{0.28\linewidth}}
  \toprule
  Model & Time enters through & Jointly fit object & Best suited regime \\
  \midrule
  DVC-smooth & basis trajectories $B(t)^\top \beta_e$ & smooth edge-parameter paths across all windows & smooth temporal evolution \\
  DVC-switch & discrete state path $F_e(t),\theta_e(t)$ & dynamic-programming-selected edge family/parameter paths across all windows & abrupt family switches; episodic interactions \\
  Win. vine control & discrete windows & independent vine per window & local refit control \\
  Reg. win. control & adjacent-window penalties & root path, family path, and drifting parameters & smoothed local-refit control \\
  DVC-latent ablation & shared latent state $z_t$ & low-rank temporal driver plus edge loadings & many edges co-evolving under a common process \\
  \bottomrule
  \end{tabular}
  }
  \end{table}

  \subsection{DVC-smooth and DVC-switch Implementation Details}
  \label{app:dvc_temporal_details}

  \paragraph{DVC-smooth.}
  DVC-smooth is used when dependence changes gradually.
  For edge $e=(i,j\mid D)$ with family $F_e$, it uses
  \[
  \eta_e(t)=B(t)^\top\beta_e,
  \qquad
  \theta_e(t)=g_{F_e}\!\left(\eta_e(t)\right).
  \]
  Here $B(t)\in\mathbb{R}^q$ is a low-dimensional time basis, $\beta_e\in\mathbb{R}^{q\times p_{F_e}}$ are coefficients shared across windows, and $g_{F_e}$ maps unconstrained latent values to valid family parameters.
  For one-parameter families, $p_{F_e}=1$.
  For Student-$t$ edges, DVC-smooth either selects a fixed degrees-of-freedom value from a grid or fits both Kendall-$\tau$/correlation and degrees of freedom as smooth trajectories, depending on the experiment.
  For a fixed family, the training objective is the edge-wise summed training NLL plus the smoothness and ridge penalties in Eq.~\eqref{eq:objective_joint}.
  The family $F_e$ is selected globally by a sequence information criterion, and after each tree level is fit, the selected $h$-functions generate pseudo-observations for the next level.
  Thus the final object is a full temporal vine rather than a collection of smoothed pairwise curves.

  \paragraph{DVC-switch.}
  DVC-switch is used when dependence changes abruptly, such as a family switch or an interaction episode that turns on and off.
  For each edge and window, candidate states $s_{e,t}=(F_{e,t},\theta_{e,t})$ are obtained by fitting candidate pair-copula families on the current pseudo-observations.
  The selected path minimizes Eq.~\eqref{eq:switching_objective}.
  Since $\mathrm{AIC}=2\mathrm{NLL}+2k$, the term $\mathrm{AIC}_{e,t}/(2N_t)$ puts the local cost on a per-observation NLL-plus-parameter-penalty scale.
  The objective is solved by dynamic programming, so DVC-switch is a temporally regularized state-selection model over locally fit candidate states, not a fully joint continuous optimization over all edge parameters.
  After each tree level, the selected time-specific copulas propagate $h$-function pseudo-observations to the next level.
  The distance $d(\theta,\theta')$ is the absolute difference between sanitized scalar parameters for one-parameter families and the mean absolute difference between transformed components for vector-valued states.
  For Student-$t$ candidate states, this includes both the correlation component and the selected degrees-of-freedom grid value after clipping to the allowed range.

  \paragraph{Windowed controls.}
  The independent Win. vine control fits a static vine separately in each window using rank pseudo-observations and AIC family selection.
  The Reg. win. control keeps the same per-window candidate fits but adds adjacent-window penalties on root changes, family switches, and parameter drift.
  These controls are intentionally strong local-refit baselines, but they do not define a single jointly estimated temporal vine.
  Where Reg. win. is reported, the adjacent-window penalties are stated in Section~\ref{app:benchmark_details}.
  
  \subsection{Evaluation and Computation Details}
  \label{app:evaluation_details}
  
  All copula models are evaluated on rank-transformed pseudo-observations.
  The reported held-out likelihoods are \emph{copula-only} likelihoods: within each time window, training and held-out observations are ranked separately using $r/(n+1)$, never pooled across train and held-out sets, yielding $u_{n,i}^{(t)}\in(0,1)$.
  This leakage-free convention is applied identically to all copula and Gaussian-copula baselines; it does create a coarser held-out grid when the test split is small, so the scores should be read as rank-scale diagnostics rather than full predictive likelihoods for raw observations.
  Thus the evaluation isolates dependence modeling rather than marginal density prediction; it should not be read as a full predictive likelihood for the raw observations.
  Because ranking is performed within windows, users should choose windows over which marginal response distributions are reasonably stable or perform a window-size sensitivity check; rapid within-window marginal drift can otherwise contaminate dependence readouts.
  Synthetic data are continuous.
  For neural spike-count summaries, we use log-transformed mean counts and add fixed-seed Gaussian jitter of scale $10^{-3}$ before ranking, which breaks ties reproducibly while leaving the population-level count summaries effectively unchanged.
  Reported NLL values are held-out copula negative log-likelihoods averaged per held-out observation unless a table or caption explicitly states a window- or session-summed quantity.
  For a full-vine model $M$, the pairwise score is evaluated by the matching 1-truncated version of the same fitted object, not by refitting a separate pairwise model:
  \[
  \widehat S_{\mathrm{pair}}^M(t)=-\mathrm{NLL}(M_{\mathrm{1trunc}},t),
  \qquad
  \widehat\Delta_{\mathrm{HO}}^M(t)=\mathrm{NLL}(M_{\mathrm{1trunc}},t)-\mathrm{NLL}(M_{\mathrm{full}},t).
  \]
  Thus DVC-switch, DVC-smooth, Win. vine, and Reg. win. can each have their own matched pair/higher-tree readout, but figures use the primary DVC readout unless a windowed control is explicitly labeled.
  Static and windowed family selection is by AIC over the candidate family set; the full parametric set is independence, Gaussian, Student-$t$, Clayton, Frank, Gumbel, and Joe, with experiment-specific subsets stated by the generating mechanism or script configuration.
  DVC-smooth selects one family per edge across time using the BIC-style sequence score described in Section~\ref{sec:joint_dynamic}, whereas DVC-switch selects a temporally penalized family/parameter path.
  For the C-vine instantiation used in the experiments, the fixed root order for DVC-smooth and DVC-switch is selected on training windows only unless supplied by an experiment: pseudo-observations are pooled across training windows, an absolute Kendall-$\tau$ dependence matrix is formed, the first root maximizes the row sum, and the remaining roots are chosen greedily by strongest remaining pooled dependence.
  This pooled-order heuristic can miss dependencies whose signed association cancels over time; when sign-changing or topology-changing dependence is expected, alternative vine orders/topologies should be cross-validated.
  The distance $d(\theta,\theta')$ in Eq.~\eqref{eq:switching_objective} is the absolute difference between sanitized scalar parameters for one-parameter families and the mean absolute difference between components for vector-valued states.
  For Student-$t$ candidate states, this includes both the correlation component and the selected degrees-of-freedom grid value after clipping to the allowed range.
  The decorrelated Allen null is computed by independently permuting each visual-area response within every session, split, and presentation-order window, preserving marginal response distributions and sample counts while destroying simultaneous cross-area dependence.
  
  \subsection{Benchmark and Hyperparameter Details}
  \label{app:benchmark_details}
  
  The main synthetic benchmark suite uses seed $2026$.
  Consequently, synthetic uncertainty summaries in the main text and tables are window-level variability for that benchmark realization, not independent dataset-seed uncertainty.
  Unless otherwise stated, each time slice is split into 80\% train and 20\% held-out observations by a deterministic seed offset for that slice.
  The multiplicative triplet uses $n=6000$ samples with $X,Y\sim\mathcal{N}(0,1)$ and $Z=XY+0.25\epsilon$.
  The pairwise-matched XOR stress test has $d=3$, $T=8$, and $N_t=3000$; the first half samples independent $U,V,W\sim\mathrm{Unif}(0,1)$ and Gaussianizes them, while the second half samples $U,V\sim\mathrm{Unif}(0,1)$, sets $W=(U+V)\bmod 1$, Gaussianizes all three coordinates, and adds $10^{-3}$ Gaussian jitter, producing near-zero pairwise correlations but deterministic triplet dependence.
  The dynamic tail-DF benchmark has $d=5$, $T=24$, $N_t=250$, fixed equicorrelation $\rho=0.6$, and Student-$t$ degrees of freedom $\nu=3$ for windows $1$--$12$ and $\nu=30$ for windows $13$--$24$.
  The tail-family switch has $d=5$, $T=24$, $N_t=250$, and fixed Kendall's $\tau=0.4$, switching from Clayton to Gumbel at $t=12$.
  The hub-switch control has $d=8$, $T=24$, $N_t=250$, hub $0\to1$ at $t=12$, and hub correlation $\rho=0.7$.
  The agent-episode benchmark has $d=6$, $T=48$, $N_t=300$, Gaussian pairwise episodes with $\rho=0.7$, and higher-order episodes generated from a Student-$t$/Clayton conditional-vine mechanism with $\rho=0.5$ and $\nu=3$.
  Its recurrent schedule is independence, pairwise, independence, higher-order, independence, mixed, independence, pairwise, higher-order, mixed, independence, with duration weights $(5,6,4,7,4,6,4,6,7,5,4)$ that round to segment lengths $(4,5,3,6,4,4,4,5,6,4,3)$ for $T=48$.
  Binary episode detection in Figure~\ref{fig:sim_episodes} thresholds each method's NLL trajectory at the independence-window mean minus $2\max(\mathrm{SD},0.01)$.
  The three-class order assignment first detects any interaction using total score above the independence-window mean plus $2\max(\mathrm{SD},0.01)$, then labels it higher-tree when $\widehat\Delta_{\mathrm{HO}}$ exceeds the independence-window mean plus $2\max(\mathrm{SD},0.005)$; mixed ground truth is collapsed with higher-tree for the reported order accuracy.
  The four-phase showcase has $d=10$, $T=60$, and $N_t=300$ observations per window, split chronologically into 85\% train and 15\% held-out samples.
  The phases occupy windows $1$--$15$, $16$--$30$, $31$--$45$, and $46$--$60$.
  Phase 1 is independent standard Gaussian noise.
  Phase 2 is a Gaussian star with root $X_0$, leaves $(X_1,X_2,X_3)$, and correlation $\rho=0.55$.
  Phase 3 keeps the same star and adds two disjoint multiplicative triplets on $(X_4,X_5,X_6)$ and $(X_7,X_8,X_9)$: for each block, independent $X,Y\sim\mathcal{N}(0,1)$ generate $Z=XY+0.10\epsilon$, after which the three coordinates are Gaussianized marginally.
  Phase 4 replaces the star/triplets with a Clayton lower-tail block on $(X_0,X_1,X_2,X_3)$ with $\theta=3.5$, leaving the remaining coordinates independent.
  Oracle information curves in Figure~\ref{fig:showcase} use analytic Gaussian-star mutual information, Monte Carlo Clayton pair information/lower-tail coefficients, and a Gauss-Hermite/Monte Carlo calculation for the multiplicative-triplet total, pair, and higher-tree components.
  
  DVC-smooth uses four time-basis functions by default: one intercept plus three Gaussian radial-basis functions on normalized time with centers $(0,0.5,1)$ and bandwidth $0.75$, smoothness penalty $\lambda_{\mathrm{smooth}}=5.0$, ridge penalty $\lambda_{\mathrm{ridge}}=10^{-3}$, L-BFGS-B optimization, and at most 80 iterations in the simulation suite unless a script states otherwise.
  For the dynamic Student-$t$ tail benchmark, DVC-smooth estimates Student-$t$ degrees of freedom with a smooth trajectory bounded in $[2.1,60]$; otherwise Student-$t$ candidate fits select from $\nu\in\{4,8,16\}$.
  DVC-switch uses local $\mathrm{AIC}/(2N_t)$ costs, family-switch penalties, and same-family parameter-drift penalties; in the main simulations $\lambda_{\mathrm{sw}}=0.08$ by default and $\lambda_{\mathrm{sw}}=0.20$ for the recurrent agent-episode benchmark, with $\lambda_{\mathrm{drift}}=0.02$ in the tail-family switch and $0$ otherwise unless stated.
  The Reg. win. control uses the same local state costs with experiment-specific adjacent-window penalties where reported: $(\lambda_{\mathrm{root}},\lambda_{\mathrm{sw}},\lambda_{\mathrm{drift}})=(0.25,0,0)$ for hub switching and $(0.10,0.20,0.10)$ for agent episodes; the dynamic tail-DF table row uses the default smoothed local-refit configuration from the benchmark script.
  The Gaussian SSM uses Fisher-$z$ pairwise correlation dynamics with observation variance $(N_t-3)^{-1}$, process variance selected from $\{0\}\cup\{10^{-6},\ldots,10^{-1}\}$ by held-out NLL, and nearest-positive-definite projection when needed.
  Graphical Lasso uses $\alpha=0.02$ with the standard scikit-learn solver, and TVGL uses $\alpha=0.02$ and $\beta=1.0$.
  We report these documented comparator settings rather than nested cross-validation for every synthetic regime; stricter tuning parity is an appropriate extension for the public benchmark release.
  The Real-NVP copula baseline uses four affine coupling blocks, hidden width 32, Adam with learning rate $10^{-3}$, batch size 64, and 60 epochs.
  MINE uses a two-hidden-layer ELU network with width 64, Adam with learning rate $10^{-3}$, batch size 128, and 100 epochs for the calibration and showcase readout.
  No rotations of Clayton, Gumbel, or Joe copulas are included in the main parametric candidate set; the parametric families therefore encode their default tail orientations.
  
  For Allen VBN, the main analysis uses 16 sessions from 8 mice with paired familiar/novel natural-image sessions:
  1064415305, 1064639378, 1087720624, 1087992708, 1090800639, 1091039902, 1092283837, 1092466205, 1093638203, 1093867806, 1099598937, 1099869737, 1115086689, 1115368723, 1119946360, and 1120251466.
  Each observation is one presentation represented by the log-transformed mean spike count in each of the five selected visual areas during the $0$--$250$ ms post-stimulus window.
  Units are filtered by minimum presence ratio $0.95$, minimum firing rate $0.1$ Hz, and at least 20 units in a retained region; the five regions with the largest retained populations are used.
  Presentation-order windows contain 120 presentations with stride 120; random split runs use 84 train and 36 held-out presentations per window, five split seeds, families $\{\mathrm{independence},\mathrm{Gaussian},\mathrm{Frank}\}$, four basis functions, $\lambda_{\mathrm{smooth}}=5.0$, $\lambda_{\mathrm{ridge}}=10^{-3}$, maximum 60 optimizer iterations, and jitter scale $10^{-3}$.
  
  \subsection{Shared Latent-State Variant}
  \label{app:latent_state}
  
  The shared latent-state variant of Section~\ref{sec:joint_dynamic} couples multiple edges through a common low-rank temporal state.
  Let $\bz_t \in \R^k$ be the latent state at time $t$ with $k \ll |E|$.
  The ablation uses
  \begin{equation}\label{eq:latent_state_dyn}
  \bz_t \approx \phi \bz_{t-1}, \qquad \eta_e(t) = w_e^\top \bz_t + b_e,
  \end{equation}
  where the approximation is enforced by a quadratic transition penalty, $\phi$ is a learned scalar autoregressive coefficient, and each edge $e$ has its own loading vector $w_e \in \R^k$ and offset $b_e \in \R$.
  The latent dimension $k$ controls how strongly edges share temporal information; $k=1$ collapses all edges to a common scalar driver, while $k = |E|$ recovers per-edge independent trajectories.
  Family selection and target Kendall-$\tau$ trajectories are first obtained from the smooth joint estimator; the latent path, loadings, offsets, and $\phi$ are then fit by joint MAP-style optimization to those target trajectories.
  No particle filtering or exact marginalization over $\bz_{1:T}$ is attempted.
  The model has the usual scale/rotation non-identifiability between latent states and edge loadings, so it is used only as a predictive low-rank ablation rather than as an interpretable latent-state model.
  This variant is most useful when many edges co-evolve under a common latent process, and is reported as an auxiliary ablation in Table~\ref{tab:dvc_variants}.
  
  \section{Gaussian Baseline Details and Extended Related Work}
  \label{app:baseline_details}
  
  This section records the Gaussian comparators used throughout the experiments.
  These baselines are not chosen to make the vine estimator win against weak alternatives, but to identify when temporal smoothing and second-order structure are already sufficient.

  \paragraph{Comparator taxonomy.}
  Table~\ref{tab:comparator_taxonomy} summarizes the diagnostic role of each comparator.
  A gap over Gaussian dynamic baselines indicates dependence not explained by time-varying second-order structure.
  A gap over the matched 1-truncated vine indicates dependence not explained even by flexible non-Gaussian first-tree copulas.
  Only the latter directly supports the higher-tree predictive claim for the chosen vine factorization.

  \begin{table}[h]
  \centering
  \caption{\textbf{Comparator taxonomy.}
  The baselines separate temporal smoothing, Gaussianity, non-Gaussian pairwise dependence, and higher-tree diagnostics.}
  \label{tab:comparator_taxonomy}
  \small
  \resizebox{\linewidth}{!}{%
  \begin{tabular}{lccccc}
  \toprule
  Comparator & Temporal coupling & Gaussian & Non-Gaussian & Higher-tree diagnostic & Family/type diagnostic \\
  \midrule
  Gaussian copula & no/windowed & yes & no & no & no \\
  Graphical Lasso & no/windowed & yes & no & no & no \\
  TVGL-style & yes & yes & no & no & no \\
  Gaussian SSM & yes & yes & no & no & no \\
  Matched 1-truncated vine & matched to model & no & yes & no & yes \\
  Full DVC & yes & no & yes & yes & yes \\
  Win. full vine & no/windowed & no & yes & yes & yes \\
  NF-copula & no/windowed & no & yes & implicit only & no \\
  MINE pairwise & no/windowed & no & yes & no & no \\
  \bottomrule
  \end{tabular}
  }
  \end{table}
  The table should be read diagnostically.
  If DVC improves over Gaussian copulas, Graphical Lasso, TVGL-style precision models, or Gaussian SSM, the data contain dependence not adequately represented by Gaussian second-order structure.
  If the full vine improves over its matched 1-truncated version, the evidence is stronger: flexible first-tree pairwise copulas are not enough, and higher-tree conditional terms improve prediction.
  Conversely, if the 1-truncated vine matches the full vine, the appropriate interpretation is non-Gaussian pairwise dependence rather than higher-tree evidence.
  
  \paragraph{Static Gaussian copula.}
  The Gaussian copula baseline Gaussianizes rank pseudo-observations within each window and fits a dense correlation matrix.
  It is a strong second-order density baseline, but it cannot represent asymmetric tails, non-elliptical pair-copula families, or conditional higher-tree dependence.
  
  \paragraph{1-truncated C-vine.}
  The 1-truncated vine keeps exactly the same first-tree edges and family-selection machinery as the corresponding full vine, but removes all higher-tree levels.
  The comparison between the full vine and this matched truncation is therefore the paper's order diagnostic: it asks whether conditional vine levels add held-out predictive likelihood beyond flexible first-tree pairwise copulas.
  
  \paragraph{Graphical Lasso.}
  Graphical Lasso is fit on Gaussianized pseudo-observations with the same per-window train/test split.
  It tests whether sparse Gaussian conditional dependence is enough, but it remains a Gaussian copula model and therefore cannot represent non-Gaussian family changes.
  
  \paragraph{TVGL-style Gaussian baseline.}
  Our TVGL-style comparator follows the standard dynamic precision-estimation template of \citet{Hallac2017}: for window-wise empirical covariance $S_t$ on Gaussianized pseudo-observations, it solves
  \[
  \min_{\Theta_t \succ 0} \sum_{t=1}^{T}\bigl[-\log\det\Theta_t + \mathrm{tr}(S_t\Theta_t)\bigr]
  + \alpha \sum_{t=1}^{T}\|\Theta_t\|_{1,\mathrm{off}}
  + \beta \sum_{t=1}^{T-1}\|\Theta_{t+1}-\Theta_t\|_F^2,
  \]
  with $\alpha=0.02$ and $\beta=1.0$ in the main benchmark suite.
  For copula evaluation, each fitted precision matrix is inverted to a covariance matrix, standardized to a unit-diagonal correlation matrix, projected to the nearest positive-definite correlation matrix when needed, and then scored as a Gaussian copula.
  
  \paragraph{Gaussian state-space baseline.}
  For smoother edge dynamics, we also track each pairwise correlation in Fisher-$z$ space with a linear Gaussian state model,
  \[
  z_{ij,t}=z_{ij,t-1}+w_{ij,t},
  \qquad
  \hat z_{ij,t}=z_{ij,t}+v_{ij,t},
  \]
  where $v_{ij,t}$ uses the standard asymptotic Fisher-$z$ observation variance $(N_t-3)^{-1}$.
  Filtered edge correlations are assembled into a positive-definite Gaussian-copula correlation matrix and evaluated by held-out copula NLL.
  
  \paragraph{Learned non-Gaussian baselines.}
  The Real-NVP/NF-copula baseline \citep{Dinh2017RealNVP,Laszkiewicz2021,Rezende2015} is trained on normal-score inputs and provides a flexible non-Gaussian density comparator without a vine-level pair/higher-tree decomposition.
  MINE is used only as a pairwise mutual-information readout in the showcase; it is not a multivariate copula model and does not test conditional higher-tree dependence.
  
  \paragraph{Why these baselines matter.}
  These Gaussian baselines are intentionally strong: they absorb temporal smoothness and sparse second-order structure, so the vine estimator only wins when non-Gaussian families or higher conditional tree levels matter.
  Additional related work on dynamic copulas, learned copula models, and copula applications in neuroscience is discussed in the bibliography and cited throughout the main text.
  Learned copula methods include implicit generative copulas \citep{Janke2021}, deep pair-copula families \citep{Ling2020}, copula normalizing flows \citep{Laszkiewicz2021}, classifier-based copula density estimation \citep{Huk2025Classifier}, diffusion/flow-based copulas with marginal-preserving dependence noising \citep{HukDamoulas2026}, and discrete copula diffusion \citep{Liu2025}\ifdvcneurips\else, as well as amortized static vine fitting with a pretrained edge estimator \citep{Safaai2026Amortized}. Neural population analyses have also examined task- and time-indexed dependence structure \citep{Safaai2025NatNeurosci}.\fi{}
  These approaches mostly improve static copula density estimation, sampling, or condition-indexed dependence readouts, whereas DVC asks how vine edges and higher-tree information contributions evolve as a jointly fitted temporal process.
  
  \section{Full Simulation Tables}
  \label{app:full_bench}
  
  The compact main figures show representative trajectories; the tables below report the same benchmark suite in aggregate form.
  For clarity, Table~\ref{tab:sim_method_comparison} reports the absolute held-out NLL for the primary estimator, the windowed and regularized vine controls, and the main Gaussian/pairwise baselines.
  The column ``Gap vs primary'' is method NLL minus the primary estimator's NLL, so positive values mean the primary DVC estimator or explicitly labeled full-vine control is better.
  
  \begin{figure}[!t]
  \centering
  \includegraphics[width=\linewidth]{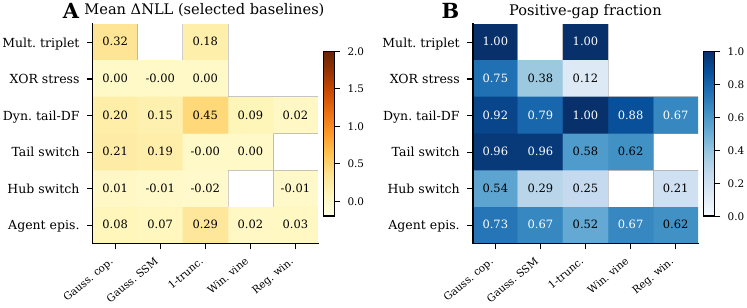}
  \caption{\textbf{Scenario$\times$baseline summary.} \textbf{(A)}~Mean held-out NLL gap for the reported estimator in each scenario (DVC-smooth or DVC-switch for temporal settings; explicitly labeled windowed full-vine controls only for static/structural diagnostics) against representative comparators, including the independent Win. vine wherever a joint DVC is available. Colorbar clipped at 2.0, multiplicative-triplet cell annotated separately. Empty cells indicate that the comparator is not applicable to that scenario (e.g.\ Win.\ vine is a control for joint-DVC settings only, and Reg.\ win.\ is reported only for the structural diagnostics that include it). \textbf{(B)}~Positive-window fraction for the same comparators, with the same applicability convention.}
  \label{fig:sim_heatmap}
  \end{figure}
  
  \begin{table}[h]
  \centering
  \caption{Primary estimator, vine controls, and baseline comparison. Values are per held-out observation. Positive gap means the scenario's primary estimator has lower held-out NLL than the row method; the positive fraction is the fraction of windows with positive gap. Scenarios whose primary estimator is ``Win. vine'' are static or structure-learning controls rather than DVC claims.}
  \label{tab:sim_method_comparison}
  \small
  \resizebox{\linewidth}{!}{\begin{tabular}{llllrrr}
\toprule
Scenario & Primary estimator & Method & Role & Mean held-out NLL & Gap vs primary & Positive-window fraction \\
\midrule
Agent episodes & DVC-switch & 1-truncated C-vine & Baseline & -0.116 & 0.286 & 0.521 \\
Agent episodes & DVC-switch & Gaussian SSM & Baseline & -0.333 & 0.070 & 0.667 \\
Agent episodes & DVC-switch & Gaussian copula & Baseline & -0.327 & 0.075 & 0.729 \\
Agent episodes & DVC-switch & Reg. win. & Control & -0.372 & 0.031 & 0.625 \\
Agent episodes & DVC-switch & Win. vine & Control & -0.380 & 0.022 & 0.667 \\
Agent episodes & DVC-switch & DVC-smooth & Joint-vine ablation & -0.165 & 0.237 & 0.938 \\
Agent episodes & DVC-switch & DVC-switch & Primary joint DVC & -0.403 & 0.000 &  \\
Dynamic tail-DF & DVC-smooth & 1-truncated C-vine & Baseline & -0.856 & 0.448 & 1.000 \\
Dynamic tail-DF & DVC-smooth & Gaussian SSM & Baseline & -1.151 & 0.153 & 0.792 \\
Dynamic tail-DF & DVC-smooth & Gaussian copula & Baseline & -1.102 & 0.202 & 0.917 \\
Dynamic tail-DF & DVC-smooth & Reg. win. & Control & -1.286 & 0.018 & 0.667 \\
Dynamic tail-DF & DVC-smooth & Win. vine & Control & -1.210 & 0.094 & 0.875 \\
Dynamic tail-DF & DVC-smooth & DVC-switch & Joint-vine ablation & -1.252 & 0.052 & 0.833 \\
Dynamic tail-DF & DVC-smooth & DVC-smooth & Primary joint DVC & -1.304 & 0.000 &  \\
XOR stress test & Win. vine & 1-truncated C-vine & Baseline & 0.001 & 0.000 & 0.125 \\
XOR stress test & Win. vine & Gaussian SSM & Baseline & 0.000 & -0.001 & 0.375 \\
XOR stress test & Win. vine & Gaussian copula & Baseline & 0.001 & 0.001 & 0.750 \\
XOR stress test & Win. vine & Win. vine & Primary full-vine control & 0.001 & 0.000 &  \\
Hub switch & Win. vine & 1-truncated C-vine & Baseline & -2.044 & -0.020 & 0.250 \\
Hub switch & Win. vine & Gaussian SSM & Baseline & -2.035 & -0.011 & 0.292 \\
Hub switch & Win. vine & Gaussian copula & Baseline & -2.016 & 0.007 & 0.542 \\
Hub switch & Win. vine & Reg. win. & Control & -2.035 & -0.011 & 0.208 \\
Hub switch & Win. vine & Win. vine & Primary full-vine control & -2.024 & 0.000 &  \\
Multiplicative triplet & Win. vine & 1-truncated C-vine & Baseline & -0.135 & 0.179 & 1.000 \\
Multiplicative triplet & Win. vine & Gaussian copula & Baseline & 0.002 & 0.316 & 1.000 \\
Multiplicative triplet & Win. vine & Win. vine & Primary full-vine control & -0.314 & 0.000 &  \\
Tail switch & DVC-switch & 1-truncated C-vine & Baseline & -0.992 & -0.002 & 0.583 \\
Tail switch & DVC-switch & Gaussian SSM & Baseline & -0.795 & 0.195 & 0.958 \\
Tail switch & DVC-switch & Gaussian copula & Baseline & -0.776 & 0.213 & 0.958 \\
Tail switch & DVC-switch & Win. vine & Control & -0.987 & 0.002 & 0.625 \\
Tail switch & DVC-switch & DVC-smooth & Joint-vine ablation & -0.790 & 0.200 & 0.917 \\
Tail switch & DVC-switch & DVC-switch & Primary joint DVC & -0.990 & 0.000 &  \\
\bottomrule
\end{tabular}
}
  \end{table}
  
  \section{Additional Scenarios: Multiplicative Triplet and Hub Switch}
  \label{app:extra_scenarios}
  
\begin{figure}[h]
  \centering
  \includegraphics[width=\linewidth]{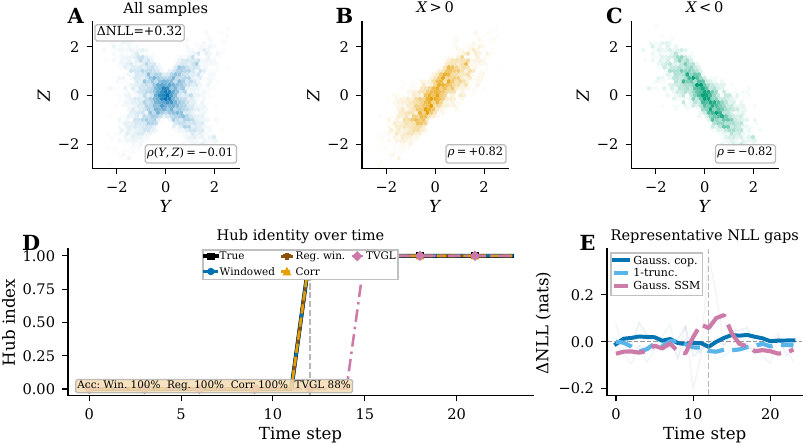}
  \caption{\textbf{Higher-order structure and structural recovery.} \textbf{(A--C)}~Multiplicative triplet shown as conditional density slices: the unconditional $Y$--$Z$ view is nearly uncorrelated, while splitting on the sign of $X$ reveals strong positive and negative dependence ($\Delta_{\text{NLL}} = +0.316$ nats over the Gaussian copula; full-vs-1-truncated gain $=+0.179$ nats). \textbf{(D)}~Hub switching: independent windowed and regularized windowed vines both achieve 100\% hub recovery and exact change-point localization, compared with a correlation-based hub tracker and TVGL. \textbf{(E)}~Representative NLL gaps for the hub-switch scenario: the near-zero 1-truncated gap confirms that this benchmark is mostly pairwise, while the Gaussian SSM excursion marks the change-point region.}
  \label{fig:sim_structure}
  \end{figure}

  \paragraph{Multiplicative triplet} ($d=3$, static): $Z = X\cdot Y + 0.25\epsilon$.
  Pairwise correlations between $Y$ and $Z$ are weak, but dependence emerges conditionally on $X$---a higher-order interaction that vine copulas capture through tree levels beyond $T_1$ while Gaussian models cannot.
  Figure~\ref{fig:sim_structure} visualizes this as raw conditional data slices rather than a model-fit panel: the unconditional $Y$--$Z$ view is nearly uncorrelated, but the sign-split views reveal strong positive and negative conditional dependence.
  The full vine achieves a NLL gap of $+0.316$ nats over the Gaussian copula and a matched full-vs-1-truncated gain of $+0.179$ nats on this benchmark, confirming a positive but modest higher-tree contribution after the Frank-copula likelihood correction.
  
  \paragraph{Hub switch} ($d=8$, $T=24$): C-vine with hub variable switching from node 0 to node 1 at $t=12$.
  This tests the windowed controls' structure-learning diagnostic: root selection (Eq.~\ref{eq:root_selection}) should track the switch.
  Both the per-window and regularized C-vine recover the hub sequence perfectly and detect the change point at $t=12$ exactly (Figure~\ref{fig:sim_structure}, panels D--E), matching correlation-based detection and outperforming TVGL ($88\%$).
  The 1-truncated vine remains near the full-vine baseline in panel~E, which is the expected outcome here: hub-switch is primarily a pairwise structural benchmark rather than a higher-tree one.
  The regularized variant also lowers held-out NLL relative to the independent windowed vine, indicating that temporal smoothing helps when the latent structure changes smoothly and only once.
  
  \section{Temporal-Vine Ablations}
  \label{app:dvc_variants}
  
  The main text uses the variant comparisons to interpret when temporal coupling helps.
  Table~\ref{tab:dvc_variants} gives the complete temporal-vine ablation summary, always measured relative to the independent windowed-vine control.
  
  \begin{table}[h]
  \centering
  \caption{Temporal-vine ablations. Positive gain means the variant has lower held-out NLL than the independent windowed-vine control.}
  \label{tab:dvc_variants}
  \small
  \resizebox{0.78\linewidth}{!}{\begin{tabular}{llrrrrr}
\toprule
Scenario & Variant & Mean gain vs Win. vine & SD & Min & Max & Positive fraction \\
\midrule
Dynamic tail-DF & DVC-switch & 0.042 & 0.064 & -0.073 & 0.175 & 0.833 \\
Dynamic tail-DF & Reg. win. & 0.076 & 0.058 & -0.030 & 0.177 & 0.875 \\
Dynamic tail-DF & DVC-smooth & 0.094 & 0.065 & -0.015 & 0.231 & 0.875 \\
Dynamic tail-DF & DVC-latent & 0.103 & 0.074 & -0.026 & 0.245 & 0.917 \\
Tail switch & DVC-switch & 0.002 & 0.020 & -0.067 & 0.041 & 0.625 \\
Tail switch & DVC-smooth & -0.198 & 0.144 & -0.520 & 0.058 & 0.125 \\
Hub switch & Reg. win. & 0.011 & 0.017 & -0.013 & 0.072 & 0.750 \\
Agent episodes & DVC-switch & 0.022 & 0.036 & -0.039 & 0.143 & 0.667 \\
Agent episodes & Reg. win. & -0.009 & 0.050 & -0.197 & 0.093 & 0.562 \\
Agent episodes & DVC-smooth & -0.215 & 0.193 & -0.711 & 0.063 & 0.083 \\
\bottomrule
\end{tabular}
}
  \end{table}
  
  \section{Real Neural Data Analyses}
  \label{app:neural_data}
  
  The neural analyses are retained as real-data evidence that vine levels beyond pairwise edges can matter in population activity.
  The main text reports the joint-DVC Allen temporal validation.
  Here we keep the Dalgleish condition-level analysis and the earlier windowed Allen cohort check as supporting material: Dalgleish is sensitive to the chosen stimulation-aligned response window, while the windowed Allen analysis verifies that the higher-tree contribution is not driven by only a few sessions or windows.
  
  \subsection{Allen Likelihood Comparisons and Interpretation}
  \label{app:allen_likelihood_interpretation}
  
  The main Allen VBN result should be interpreted as a null-calibrated temporal dependence signal rather than as uniform likelihood dominance over all Gaussian temporal baselines.
  Across five random held-out splits, joint DVC achieves a modest positive temporal gain relative to a constant sequence-wide vine:
  \[
  +0.016\pm0.016
  \]
  nats per held-out presentation, with positive session-split mean in $68/80$ fits.
  It also achieves a positive gap relative to the independent Win. vine control:
  \[
  +0.084\pm0.040
  \]
  nats per held-out presentation, with positive gaps in $78/80$ fits.
  This comparison is partly an overfitting check because the Win. vine refits each window independently.
  
  The Gaussian state-space model remains slightly better in held-out likelihood on this neural representation.
  Using the SSM-minus-DVC gap convention, the mean gap is
  \[
  -0.010\pm0.022
  \]
  nats per held-out presentation, with DVC positive in $23/80$ fits.
  Thus likelihood alone does not require a non-Gaussian temporal copula model for this particular Allen representation.
  The main positive result is the decomposition and null calibration:
  \[
  \widehat S_{\mathrm{total}}=1.020\pm0.702,
  \qquad
  \widehat S_{\mathrm{pair}}=0.854\pm0.672,
  \qquad
  \widehat\Delta_{\mathrm{HO}}=0.166\pm0.086
  \]
  nats per held-out presentation, with positive higher-tree contribution in all $80/80$ session-split fits.

  Finally, a decorrelated null calibrates the signal.
  We independently permute each visual-area response within every train/test split and presentation-order window, preserving marginal response distributions and temporal sampling while destroying simultaneous cross-area interactions.
  Under this null, joint-DVC temporal gain and both pairwise and higher-tree score estimates collapse to the independence floor:
  \[
  0.000
  \]
  nats, with higher-tree positive in $0/80$ fits.
  We therefore interpret Allen as evidence for a modest but reproducible time-indexed higher-tree dependence signal that disappears when simultaneous cross-area dependence is destroyed, while the Gaussian SSM comparison prevents overclaiming a likelihood win over all Gaussian temporal models.
  
  \subsection{Dalgleish Preprocessing Robustness}
  \label{app:dalgleish_robustness}
  
  The Dalgleish supporting result uses the public two-photon photostimulation dataset of \citet{Dalgleish2020} and its figshare release \citep{DalgleishFigshare}.
  It depends on the stimulation-aligned response window (the \texttt{stim\_post} backbone) in which the post-stimulation evaluation window is $[1.0\,\mathrm{s}, 2.0\,\mathrm{s}]$ relative to photostimulation onset.
  Under this backbone, at the session-aggregated level, the full vine achieves positive held-out NLL gaps relative to Graphical Lasso ($+0.102$ nats), Gaussian SSM ($+0.093$), the static Gaussian copula ($+0.073$), and the 1-truncated vine ($+0.056$), so the matched higher-tree held-out gain is $\widehat\Delta_{\mathrm{HO}} = +0.056$ nats.
  
  \begin{figure}[h]
  \centering
  \includegraphics[width=\linewidth]{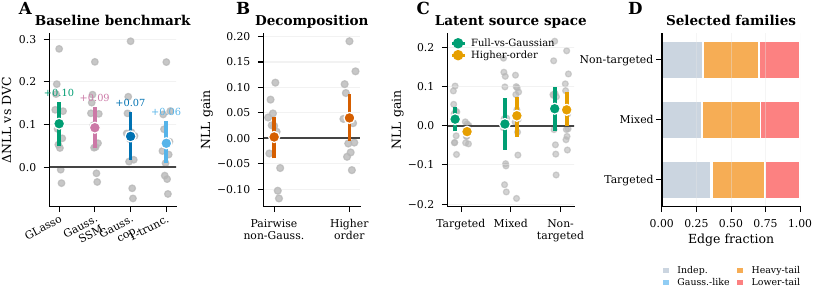}
  \caption{\textbf{Dalgleish stimulation-aligned latent-state analysis.} \textbf{(A)}~Session-level baseline comparison for the selected non-targeted latent variant: the full vine has positive gaps relative to Graphical Lasso, Gaussian SSM, Gaussian copula, and the 1-truncated vine. \textbf{(B)}~The gain decomposes into a near-zero pairwise-flexible component and an estimated higher-tree contribution that is positive on average, supporting but not by itself proving dependence beyond pairwise structure alone. \textbf{(C)}~The strongest signal appears in the non-targeted latent population, whose higher-tree gain exceeds the targeted-only space. \textbf{(D)}~Selected pair-copula families are dominated by heavy-tailed elliptical and lower-tail-asymmetric structure, with little Gaussian-like usage. The figure reports the reproduced \texttt{stim\_post} backbone and should be interpreted with that explicit preprocessing choice in mind.}
  \label{fig:real_dalgleish}
  \end{figure}
  
  An alternative backbone---the \texttt{delayed\_post} window $[0.7\,\mathrm{s}, 1.4\,\mathrm{s}]$---produces materially weaker and more heterogeneous results under the same analysis pipeline.
  Under a \texttt{stim\_mean} configuration with $d=4$ latent dimensions, the full vine beat the static Gaussian copula in only $50.6\%$ of session-dose-repeat slices (mean $\Delta_{\mathrm{NLL}}$ against Gaussian $-0.024$ nats), and the mean higher-tree gain was $-0.019$ nats (median $+0.012$), with dose-level behavior ranging from $+0.066$ at low doses to $-0.079$ at the highest dose.
  We therefore treat the Dalgleish result as a supportive case that a higher-tree signal is detectable in a real neural population dataset, not as a preprocessing-invariant benchmark.
  The sensitivity reflects a known issue: the relative timing of photostimulation-induced activity and the baseline window changes which latent components dominate the principal component analysis (PCA) projection, and conditional-on-latent copula structure differs between the two.
  A fully preprocessing-robust real-data benchmark---e.g., jointly modeled across windows with an explicit stimulus kernel---is a natural follow-up that is out of scope for the current paper.
  Therefore, the current Dalgleish result should be read as a static or condition-level real-data copula result, not as a fitted temporal-DVC trajectory.
  
  \subsection{Additional Windowed Allen Visual Behavior Neuropixels Cohort Check}
  \label{app:allen_vbn}
  
  As a cohort-level real-data check complementary to the main joint-DVC Allen analysis, we also ran the independent windowed full-vine control on the same 16 Allen Visual Behavior Neuropixels sessions spanning 8 mice and paired familiar/novel natural-image conditions \citep{AllenVBNData}.
  Each session used five visual areas and roughly 4.6k selected presentations.
  The full vine beat the 1-truncated vine in every session: the mean session-level full-vs-1-truncated held-out gap was $34.6 \pm 7.2$ nats (mean $\pm$ SD across sessions), with positive-gap fraction $1.0$ in all 16 sessions.
  Familiar and novel sessions had similar mean gaps ($33.9 \pm 5.2$ versus $35.3 \pm 8.4$ nats), and the paired mouse-level novel-minus-familiar change was $1.4 \pm 7.0$ nats.
  This analysis is windowed over presentation order rather than a single static fit or a jointly fitted DVC trajectory: with 120-presentation windows, each session contributes 38 temporal windows.
  The higher-tree gap varied within sessions (mean within-session SD $13.2$ nats across presentation-order windows), so the Allen cohort provides an additional time-indexed dependence check.
  Across all $608$ temporal windows, the full-vs-1-truncated gap was positive in every window, making the cohort result easier to interpret as a persistent higher-tree contribution rather than a few extreme sessions.
  However, the windowed gap had weak average association with image-change fraction ($r=-0.085 \pm 0.152$), mean absolute pairwise correlation ($r=-0.123 \pm 0.227$), and normalized presentation time ($r=-0.010 \pm 0.261$).
  Thus the signal is not well explained by pairwise correlation magnitude or a simple monotone session drift.
  This supports the main claim that vine levels beyond pairwise edges carry measurable, time-indexed dependence in real neural population data, while still treating the cohort as a validation check rather than a jointly modeled temporal-DVC experiment or a fully controlled mechanistic result.
  
  \begin{figure}[h]
  \centering
  \includegraphics[width=\linewidth]{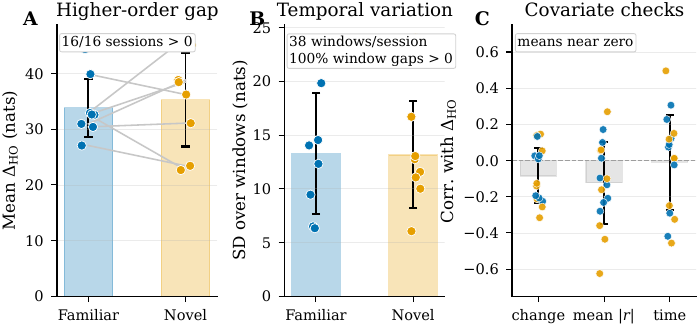}%
  \caption{\textbf{Allen Visual Behavior Neuropixels (VBN) cohort validation.} (A) Session-level mean higher-tree held-out gap $\widehat\Delta_{\mathrm{HO}}=\mathrm{NLL}_{\text{1-trunc}}-\mathrm{NLL}_{\mathrm{full}}$ for paired familiar and novel sessions from 8 mice. Bars show mean $\pm$ SD, dots are sessions, and gray lines connect paired mice. Values are session-summed nats across held-out presentations/windows, unlike the per-held-out-presentation values in Figure~\ref{fig:allen_joint_dvc}. (B) Within-session temporal variation of the same windowed gap, summarized by the SD across 38 presentation windows per session. (C) Session-level correlations between the windowed gap and image-change fraction, mean absolute pairwise correlation, and presentation time. Positive gaps indicate held-out improvement from higher vine levels beyond the pairwise-only truncation.}
  \label{fig:allen_vbn_cohort}
  \end{figure}
  
  \section{Runtime and Scaling}
  \label{app:runtime}
  
  Joint DVC changes the fitted object: it estimates temporal edge trajectories or state paths rather than a sequence of unrelated static vines.
  Table~\ref{tab:runtime} reports both wall-clock fit time and the number of edge-level fit objects for the windowed and smooth joint C-vine estimators as a function of $d$ (number of variables) and $T$ (number of time windows).
  A windowed vine refits $T\,d(d-1)/2$ pair-copula edges, whereas smooth joint DVC fits $d(d-1)/2$ temporal edge trajectories; this is a $T$-fold compression in the fitted dependence objects.
  In the current Python implementation, this compression does not yet translate into faster wall-clock time on the small single-CPU benchmarks because each trajectory is optimized with generic, unvectorized routines.
  Joint temporal objectives are also nonconvex, and sequential $h$-function propagation can carry lower-tree estimation error into higher trees; the oracle consistency statement in Appendix~\ref{app:proofs} assumes this propagation error is controlled rather than claiming it away.
  Thus the present result should be read as evidence for temporal model compression and a clear optimization target, not as a measured speedup claim.
  
  \begin{table}[h]
  \centering
  \caption{Runtime scaling for windowed and smooth joint C-vine estimators. ``Edge fits'' counts the number of edge-level objects fit by the method: separate per-window pair-copulas for Win. vine versus one temporal trajectory per edge for joint DVC. Total time is the sequential wall-clock time for the complete $T$-window sequence, averaged over $2$ repeats on one CPU core; time/window is the total divided by $T$.}
  \label{tab:runtime}
  \small
  \resizebox{\linewidth}{!}{\begin{tabular}{rrlrrrr}
\toprule
d & T & variant & edge fits & compression & total time (s) & time / window (s) \\
\midrule
3 & 12 & Windowed & 36 & 1$\times$ & 1.07 & 0.090 \\
3 & 12 & Joint dynamic & 3 & 12$\times$ & 1.54 & 0.128 \\
3 & 24 & Windowed & 72 & 1$\times$ & 1.91 & 0.080 \\
3 & 24 & Joint dynamic & 3 & 24$\times$ & 3.61 & 0.150 \\
5 & 12 & Windowed & 120 & 1$\times$ & 3.60 & 0.300 \\
5 & 12 & Joint dynamic & 10 & 12$\times$ & 5.10 & 0.425 \\
5 & 24 & Windowed & 240 & 1$\times$ & 7.30 & 0.304 \\
5 & 24 & Joint dynamic & 10 & 24$\times$ & 9.92 & 0.413 \\
8 & 12 & Windowed & 336 & 1$\times$ & 10.45 & 0.871 \\
8 & 12 & Joint dynamic & 28 & 12$\times$ & 14.38 & 1.198 \\
8 & 24 & Windowed & 672 & 1$\times$ & 21.34 & 0.889 \\
8 & 24 & Joint dynamic & 28 & 24$\times$ & 26.49 & 1.104 \\
\bottomrule
\end{tabular}
}
  \end{table}
  
  For a more formal comparison, let
  \[
  n_E=\frac{d(d-1)}{2}
  \]
  be the number of full-vine pair-copula edges, let $K$ be the number of candidate pair-copula families, let $q$ be the number of temporal basis functions in smooth joint DVC, and let $I_{\mathrm{win}}$ and $I_{\mathrm{joint}}$ denote typical optimizer iterations for static and joint edge fits.
  The leading terms in Table~\ref{tab:complexity} suppress rank transforms $O(TdN\log N)$, h-function propagation $O(Tn_EN)$, and fixed-order construction costs, which are lower-order terms when family fitting dominates.
  
  \begin{table}[h]
  \centering
  \caption{\textbf{Analytical fitting complexity.}
  All expressions are for fitting the complete $T$-window sequence with $N$ samples per window and dimension $d$.
  The key distinction is between repeated local fitting and temporal parameterization: smooth joint DVC still evaluates all $TN$ samples, but it fits one trajectory per edge rather than $T$ unrelated edge models.}
  \label{tab:complexity}
  \small
  \resizebox{\linewidth}{!}{%
  \begin{tabular}{p{0.20\linewidth}p{0.34\linewidth}p{0.18\linewidth}p{0.22\linewidth}}
  \toprule
  Method & Leading fit complexity & Fitted edge objects & Interpretation \\
  \midrule
  Gaussian SSM
  & $O\!\left(TNd^2 + Td^3\right)$
  & $O(d^2)$ filtered correlations
  & Strong when dependence is Gaussian and pairwise. \\
  Win. full vine
  & $O\!\left(Tn_EK I_{\mathrm{win}}N\right)$
  & $Tn_E$
  & Refits a static full vine independently in every window; easy to parallelize. \\
  Reg. win.
  & $O\!\left(Tn_EK I_{\mathrm{win}}N + Tn_EK^2\right)$
  & $Tn_E$
  & Adds temporal penalties/path costs but still relies on local per-window candidate fits. \\
  DVC-switch
  & $O\!\left(Tn_EK I_{\mathrm{win}}N + Tn_EK^2\right)$
  & $n_E$ temporal state paths, each length $T$
  & Jointly selects family/state paths; same leading local-candidate cost as windowed fitting. \\
  DVC-smooth
  & current: $O\!\left(n_EK I_{\mathrm{joint}}qTN\right)$; with analytic gradients: $O\!\left(n_EK I_{\mathrm{joint}}TN\right)$
  & $n_E$ temporal trajectories, $O(n_Eq)$ coefficients
  & Replaces $T$ static edge fits by one smooth trajectory per edge. \\
  \bottomrule
  \end{tabular}
  }
  \end{table}
  
  These expressions make clear when a wall-clock win is expected.
  Ignoring the common $n_EK$ factor, write the sequential windowed cost as
  \[
  C_{\mathrm{win}}(T,N)\approx T\left(h_{\mathrm{win}}+a_{\mathrm{win}} I_{\mathrm{win}}N\right)
  \]
  and the current smooth joint cost as
  \[
  C_{\mathrm{joint}}(T,N)\approx h_{\mathrm{joint}}+a_{\mathrm{joint}} I_{\mathrm{joint}}qTN,
  \]
  where $h_{\mathrm{win}}$ and $h_{\mathrm{joint}}$ are per-optimization overheads and $a_{\mathrm{win}},a_{\mathrm{joint}}$ are per-sample likelihood constants.
  Smooth joint DVC is faster when
  \[
  T >
  \frac{h_{\mathrm{joint}}}
  {h_{\mathrm{win}}+a_{\mathrm{win}}I_{\mathrm{win}}N-a_{\mathrm{joint}}I_{\mathrm{joint}}qN},
  \]
  provided the denominator is positive.
  Thus the likely wall-clock win regime is many short windows (large $T$, modest $N$), small $q$, and an optimized vectorized or analytic-gradient implementation that makes the joint per-sample constant competitive.
  If the denominator is nonpositive, as in the small CPU grids in Table~\ref{tab:runtime}, there is no serial wall-clock crossover even though the model-object and parameter counts are much smaller.
  
  \section{MINE Calibration}
  \label{app:mine_calibration}
  
  To verify that the MINE baseline \citep{Belghazi2018} used in Section~\ref{sec:results} returns calibrated mutual-information estimates at the sample sizes of our showcase, we run a standalone bivariate-Gaussian sweep.
  For $\rho \in \{0.0, 0.2, 0.4, 0.6, 0.8\}$, we draw $n=500$ samples from a zero-mean bivariate Gaussian with correlation $\rho$, run MINE for 100 epochs with the same hyperparameters used in the main-text showcase, and average over 3 independent seeds.
  Table~\ref{tab:mine_calibration} reports the MINE estimate against the analytic MI $-\tfrac{1}{2}\log(1 - \rho^2)$.
  
  \begin{table}[h]
  \centering
  \caption{MINE calibration on bivariate Gaussians ($n=500$, 3 seeds, same hyperparameters as the showcase). MINE consistently underestimates the analytic MI by $7$--$10\%$ at moderate-to-large $\rho$, a known finite-sample bias of the Donsker--Varadhan bound; the independent case is near zero.}
  \label{tab:mine_calibration}
  \small
  \resizebox{0.55\linewidth}{!}{\begin{tabular}{rrrr}
\toprule
$\rho$ & true MI (nats) & MINE mean (nats) & MINE std (nats) \\
\midrule
$0.00$ & $+0.000$ & $+0.002$ & $0.001$ \\
$0.20$ & $+0.020$ & $+0.019$ & $0.007$ \\
$0.40$ & $+0.087$ & $+0.079$ & $0.015$ \\
$0.60$ & $+0.223$ & $+0.204$ & $0.027$ \\
$0.80$ & $+0.511$ & $+0.473$ & $0.040$ \\
\bottomrule
\end{tabular}
}
  \end{table}
  
  A regression test (\texttt{tests/test\_mine\_baseline.py}) pins these calibration properties and runs under the standard test suite.
  
  \section{Asset Licenses}
  \label{app:licenses}
  
  Table~\ref{tab:licenses} consolidates the external assets used in this work and their licenses. All assets are used within the terms of their respective licenses.
  The final build environment used Python 3.10.13, NumPy 1.26.4, SciPy 1.15.3, PyTorch 2.9.1, matplotlib 3.10.6, and pyvinecopulib 0.7.5.
  
  \begin{table}[h]
  \centering
  \caption{Consolidated license table for external assets used in this work.}
  \label{tab:licenses}
  \small
  \resizebox{\linewidth}{!}{%
  \begin{tabular}{lll}
  \toprule
  Asset & Source & License \\
  \midrule
  DVC implementation (this work)          & \dvcCodeAssetSource{} & MIT \\
  PyTorch                                 & \url{https://pytorch.org}               & BSD-3-Clause \\
  NumPy                                   & \url{https://numpy.org}                 & BSD-3-Clause \\
  SciPy                                   & \url{https://scipy.org}                 & BSD-3-Clause \\
  matplotlib                              & \url{https://matplotlib.org}            & Matplotlib License (PSF-based) \\
  pyvinecopulib \citep{pyvinecopulib}     & \url{https://github.com/vinecopulib/pyvinecopulib} & MIT \\
  VineCopula (R) \citep{VineCopulaR}      & CRAN                                    & GPL-2+ \\
  VineCopulas (Python) \citep{VineCopulasPython2024} & pip                          & MIT \\
  Dalgleish photostimulation dataset \citep{Dalgleish2020,DalgleishFigshare} & figshare DOI \texttt{10.6084/m9.figshare.13114247} & as stated on figshare record \\
  Allen Visual Behavior Neuropixels \citep{AllenVBNData} & DANDI DOI \texttt{10.48324/dandi.000713/0.240702.1725} & Allen Institute Terms of Use \\
  \bottomrule
  \end{tabular}
  }
  \end{table}
  \FloatBarrier

\end{document}